\documentclass[letterpaper, 10 pt, conference]{ieeeconf}  % Comment this line out if you need a4paper

\IEEEoverridecommandlockouts                          
\usepackage{graphics} % for pdf, bitmapped graphics files
\usepackage{epsfig} % for postscript graphics files
\usepackage{mathptmx} % assumes new font selection scheme installed
\usepackage{times} % assumes new font selection scheme installed
\usepackage{amsmath} % assumes amsmath package installed
\usepackage{amssymb}  % assumes amsmath package installed

\newcommand{\etal}{\textit{et al.}}
\usepackage{xcolor}
\usepackage{hyperref}
\usepackage{float}
\usepackage{subfig}
\usepackage{multirow}
\usepackage{caption}
\title{\LARGE \bf
Indoor Scene Recognition in 3D
}

\author{Shengyu Huang, Mikhail Usvyatsov and Konrad Schindler$^{\dagger}$% <-this % stops a space
\thanks{$^{\dagger}$The authors are with the Photogrammetry and Remote Sensing group, ETH Zurich, 8093 Zurich, Switzerland (email: shenhuan@student.ethz.ch, mikhail.usvyatsov@geod.baug.ethz.ch, schindler@ethz.ch)}%
}

\begin{document}

\maketitle
\thispagestyle{empty}
\pagestyle{empty}

%%%%%%%%%%%%%%%%%%%%%%%%%%%%%%%%%%%%%%%%%%%%%%%%%%%%%%%%%%%%%%%%%%%%%%%%%%%%%%%%
\begin{abstract}
  Recognising in what type of environment one is located is an important perception task. For instance, for a robot operating indoors it is helpful to be aware whether it is in a kitchen, a hallway or a bedroom. Existing approaches attempt to classify the scene based on 2D images or 2.5D range images. Here, we study scene recognition from 3D point cloud (or voxel) data, and show that it greatly outperforms methods based on 2D birds-eye views. Moreover, we advocate multi-task learning as a way to improve scene recognition, building on the fact that the scene type is highly correlated with the objects in the scene, and therefore with its semantic segmentation into different object classes. In a series of ablation studies, we show that successful scene recognition is not just the recognition of individual objects unique to some scene type (such as a bathtub), but depends
  on several different cues, including coarse 3D geometry, colour, and the (implicit) distribution of object categories. Moreover, we demonstrate that surprisingly sparse 3D data is sufficient to classify indoor scenes with good accuracy.
  \end{abstract}

%%%%%%%%%%%%%%%%%%%%%%%%%%%%%%%%%%%%%%%%%%%%%%%%%%%%%%%%%%%%%%%%%%%%%%%%%%%%%%%%
\section{INTRODUCTION}
An autonomous agent's behaviour strongly depends on what type of environment they are in: the knowledge that we are, say, in a kitchen and not in a bathroom changes what set of possible actions we consider, what objects we expect in the environment and where we search for them, and even how we navigate and move around.
The fact that the type of environment is such a strong prior for elementary behaviours suggests that an autonomous robot should have the ability to determine what environment it is in. I.e., it should have the ability to perform \emph{scene recognition} based on its sensory input.

Here we are concerned with indoor scene recognition, where the set of scenes is a list of different room types defined mostly by the function they serve, such as kitchen, bedroom, hallway, etc.%
\footnote{%
Note that indoor scene recognition is also useful for a robot that operates in a more diverse set of environments, since it is efficient to follow a coarse-to-fine hierarchy, e.g., first determine whether one is outdoors, in a residential building or in an industrial facility; then switch to dedicated, more fine-grained scene
categories.} %
Especially indoors, the scene type depends on a variety of cues, including for instance the global geometry (e.g., corridors), the presence of specific objects (e.g., a washing machine), and the relative placement of objects (e.g., chairs around a dining table vs. a single chair at a desk).
In most cases it is not obvious which visual properties of a scene are most discriminative -- is it the size? the 3D shape? specific textures? individual objects?

\begin{figure}[t!]
\begin{center}
   \includegraphics[width=1.0\linewidth]{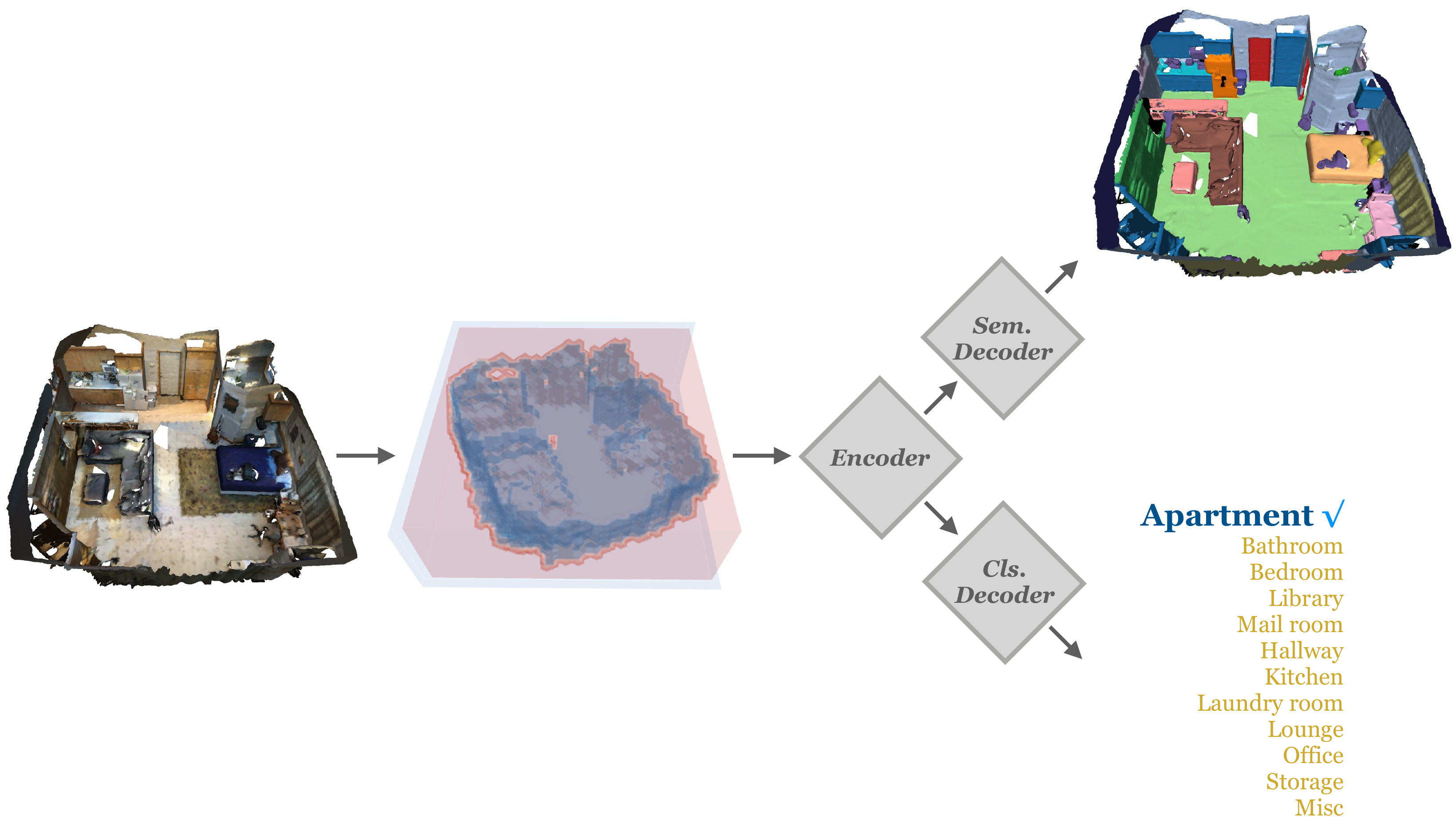}
\end{center}
   \caption{We demonstrate improved classification of indoor scenes by working with subsampled or voxelised 3D point clouds. Multi-task learning together with semantic segmentation further boosts performance.}
\label{fig: overview}
\end{figure}

Scene recognition was first explored as a 2D image classification problem~\cite{guerin00prl,xiao2010cvpr-sun}.
For indoor scenes, it has also been suggested to operate in 2.5D, using RGB-D images from range cameras or stereo rigs~\cite{gupta2015indoor,banica2015second}.
On the contrary, to our knowledge there have not been any attempts to perform scene classification in 3D point cloud or voxel data. This is somewhat surprising, given that the scene, and in most cases also (part of) the sensory input of a robot, are 3-dimensional.
We attribute this gap to two reasons. First, machine learning is computationally more expensive in 3D, in particular efficient deep learning architectures for 3D point cloud or voxel data have only been developed in the last three years. Second, other than for 2D images and 2.5D range scans, there are no suitable public datasets. So far the only available benchmark we are aware of is ScanNet~\cite{dai2017scannet} with a modest 1613 scans covering 21 different scene types, a small fraction of today's 2D image~\cite{deng2009imagenet}, range image~\cite{xiao2013sun3d}, and video~\cite{kay2017kinetics} datasets.

In this paper we take a first step to close the gap and study scene recognition based on 3D data.
An important observation in that context is the following: What is ``small'' about the datasets is the number of scenes, and consequently the information content of the ground truth, $\approx\,$1000 integer labels.
Whereas the dataset, designed also for semantic segmentation, has much richer per-point labels as well. We argue that the pointwise labels constitute valuable, and much more plentiful side information to guide the learning towards semantically meaningful features.
To that end, we employ multi-task learning, where semantic segmentation is learned as an auxiliary task during training, with a shared encoder for both the segmentation and scene classification tasks, see Fig.~\ref{fig: overview}.
For both the encoder and the semantic segmentation decoder we use sparse 3D learning architectures that allow for efficient processing on standard hardware, since 3D scenes have a high degree of sparsity (as most of the volume is empty).
Our 3D encoder-decoder network sets a greatly improved state of the art for scene type classification on ScanNet, compared to previous 2D methods. Moreover, we find that the multitask learning boosts the classification performance even further, to an overall accuracy of 90.3\%.
We also perform ablation studies to disentangle the influence of geometry, object semantics and colour, and the influence of 3D point density.
We find that the different cues are somewhat complementary. Without object information, scene geometry does the heavy lifting, although using also colour can improve the classification performance, in some cases significantly.
On the other hand, the distribution of object classes in the scene, without any geometry, is also a surprisingly good predictor.
Moreover, our experiments indicate that scene recognition is not simply the recognition of single, distinctive objects, as one might suspect. For most scene types, removing individual object classes has only a small effect.
Somewhat unexpectedly, classification performance drops only a little if one severely downsamples the input point cloud: representing an entire room with 1024 randomly sampled points is sufficient to all but match the performance on the full point cloud.
That the scene type can indeed be predicted quite well from a rough ``overall glance'' at the environment supports our strategy to infer it early on, so as to customise subsequent, more fine-grained perception and action tasks.

\section{Related work}
%In this section, we first introduce scene classification in 2D and 2.5D, then we introduce two 3D deep models: Pointnets that operate on points in continuous space and sparse 3D convolution that operates on voxel grids, finally we introduce multi-task learning in 3D.

\subsection{Scene recognition in 2D}
Scene recognition from 2D images has been investigated since at least~\cite{guerin00prl}, and several datasets are available, including Scene15~\cite{lazebnik2006beyond}, MIT Indoor67~\cite{quattoni2009recognizing}, SUN397~\cite{xiao2010cvpr-sun} and Places476~\cite{zhou2014learning}.  
Early works use handcrafted features to capture discriminative scene characteristics. 
Oliva \etal~\cite{oliva2001modeling} propose a set of perceptual dimensions named spatial envelope (e.g.\ naturalness, openness, roughness, expansion, ruggedness) that represent the dominant spatial structure of a scene.
Gokalp \etal~\cite{gokalp2007scene} cluster the regions of an image partitioning to obtain a codebook of region types, then classify based on the bag of individual and paired region types.
Bosch \etal~\cite{bosch2008scene}  apply probabilistic Latent Semantic Analysis (pLSA) to discover “topics” from a bag of visual words and train a multi-way classifier on the topic distribution vectors. 
Li \etal~\cite{li2010object} represent an image as a scale-invariant response map of a large number of pre-trained generic object detectors named "Object Bank". 

Recent methods employ deep convolutional neural network (CNN) features.
Zhou \etal~\cite{zhou2014learning} introduce a bigger scene-centric dataset, Places476, and use a CNN to learn features for scene recognition.
Herranz \etal~\cite{Herranz_2016_CVPR} efficiently combine scene-centric Places-CNNs and object-centric ImageNet-CNNs by taking the feature scales into consideration.
Cheng \etal~\cite{cheng2018scene} rely on object detectors and represent images via the occurrence probabilities of discriminative objects contained in them. 
%

% 2010, SUN397~\cite{xiao2010cvpr-sun} RGB dataset for scene recognition. We measure human scene classiﬁcation performance on the SUN database and compare this with computational methods.

\subsection{Scene recognition in 2.5D}
Especially in indoor scenes, depth can be a valuable additional cue for scene recognition.
Silberman \etal~\cite{silberman2012indoor} introduce the view-centric NYU dataset of RGB-D images and show that the combination of depth and intensity significantly improves scene understanding. 
%
%Xiao \etal~\cite{xiao2013sun3d} introduce place-centric SUN3D dataset which contains 254 scenes over 11 scene types. 
Song \etal~\cite{song2015sun} introduce the larger SUN RGB-D dataset to support data-hungry learning methods. 

Like in the case of 2D images, early works use handcrafted features~\cite{gupta2015indoor,banica2015second} while the current state-of-the-art are deep features from neural networks.
Wang \etal~\cite{wang2016modality} encode distributions of CNN features computed from RGB-D data using the Fisher vector representation, to allow for greater spatial flexibility. 
Zhu \etal~\cite{zhu2016discriminative} first extract CNN features from RGB and depth separately, then fuse them through a multi-modal layer which considers both inter- and intra-modality correlations and regularises the learned features.
Song \etal~\cite{song2018learning} address the problem of limited range of depth sensors by learning from RGB-D videos that contain comprehensive, accumulated depth information, using convolutional RNNs. %they combine convolutional and recurrent neural networks (RNNs) that are trained in three steps with increasingly complex data to learn effective features. 
Du \etal~\cite{du2019translate} formulate modality-specific scene recognition and cross-modal translation under a multi-task learning setting, explicitly regularizing the recognition task by training a joint encoder. 
%
%~\cite{gupta2016cross} propose a cross-modal distillation approach where learning of depth ﬁlters is guided by the high-level RGB features obtained from the paired RGB image the greater spatial variability in scene images typically meant that the standard full-image CNN features are suboptimal for scene classification

\subsection{Deep learning in 3D}
3D data can be represented by multiple images with different viewing directions~\cite{su2015multi,kalogerakis20173d}, as voxel grids~\cite{maturana2015voxnet,qi2016volumetric}, point clouds~\cite{qi2017pointnet,qi2017pointnet++}, or meshes~\cite{defferrard2016convolutional,gkioxari2019mesh}. Here, we limit ourselves to point clouds and voxel grids.

Pointnet~\cite{qi2017pointnet} was a seminal work for deep learning directly from point clouds. The basic idea is to apply a shared bank of MLPs to the coordinates and attributes of each individual point, as approximations of point kernels to extract point-wise features, then use global max-pooling to abstract to a global representation in a permutation-invariant manner. By design, Pointnet does not capture local structures induced by the metric space the points live in, making it difficult to deal with geometric detail. To overcome this limitation, the follow-up version Pointnet++~\cite{qi2017pointnet++} applies Pointnet recursively on a nested partitioning of the input point set, to hierarchically aggregate local information into a compact and fine-grained representation. 

Voxels are a straight-forward 3D generalisation of pixels, but point clouds from 3D sensors are sparse by nature. Instead of applying 3D convolutions on the full volumetric occupancy grid~\cite{maturana2015voxnet}, sparse CNNs store the non-empty voxels as sparse tensors and only performs convolutions on such sparse coordinate lists. Graham \etal~\cite{graham2014spatially} introduce a CNN which takes sparsity into account, but is limited to small resolution ($80^3$ voxels in their experiments) due to the decrease in sparsity after repeated convolution. To deal with the dilation of non-zero activations, Graham \etal ~\cite{graham2017submanifold} advocate the strategy to store the convolution output only at occupied voxels. Hackel \etal ~\cite{hackel2018inference} explore feature sparsity by selecting only a fixed number of the highest activations. Choy \etal~\cite{choy20194d} introduce MinkowskiEngine, an open-source auto-differentiation library that extends~\cite{graham2017submanifold} to 4D spatio-temporal perception. It also proposes hybrid kernels with predefined sparsity patterns to mitigate the exponential increase of parameters. 
% One disadvantage of Pointnets is that they require a fixed number of points as input, as this is usually not the case in real-world scenarios. Sub-sampling like Farthest Point Sampling~\cite{qi2017pointnet++} is usually required to deal with such a problem. However, this will introduce redundant information(up-sampling) or lose fine-grained information(downsampling). 

% Instead of fixing the number of input points, sparse convolution requires a fixed voxel resolution used to discretize the point cloud. Though sub-sampling is still applied via uniform sampling during discretisation, the resolution remains the same across the scan.

\subsection{Multi-task learning in 3D}
Deep models contain millions of parameters and require a large number of samples to supervise the learning. However, it is not always possible to collect large training sets, especially in applications where data cannot be gleaned from the internet and/or cannot be annotated by non-experts, like for example medical image analysis~\cite{zhang2017survey}. Also 3D scene recognition falls into the category of tasks for which massive supervision is difficult, due to the sheer effort of collecting 3D scans of many environments (where each scan only constitutes a single training example). In such cases, multi-task learning (MTL)~\cite{caruana1997multitask} is a sensible way to exploit annotations for related tasks and transfer useful information to tasks with limited training data.

Multi-task learning aims to improve the performance of multiple related learning tasks by sharing information among them. Several authors jointly learn semantic segmentation and instance segmentation with a shared latent representation. For example, Wang \etal~\cite{wang2019associatively} jointly optimise semantic labeling and an embedding that discriminates individual object instances. A related approach, Pham \etal~\cite{pham2019jsis3d} fuses semantic labels and instance embeddings into a conditional random field and makes the final predictions with variational inference. Lahoud \etal~\cite{lahoud20193d} also tackle semantic segmentation and instance segmentation, with multi-task metric learning.
To our knowledge, our work is the first to combine 3D scene recognition and semantic segmentation in the multi-task framework.

\section{Method}

\subsection{Basic 3D learning framework}
We treat scene recognition as a supervised classification problem and solve it with a neural network. The network consists of an encoder that transforms the input scene $S$ into a feature representation $Z$, followed by a classification head to compute the class-conditional likelihood $Y$.
For the encoder, we explore two different options: networks that work with a subsampled version of the original point cloud as input (Pointnet~\cite{qi2017pointnet}, Pointnet++~\cite{qi2017pointnet++}, DGCNN~\cite{wang2019dynamic}),  and networks that work with a sparse voxel grid derived from the input points (Resnet14~\cite{he2016deep} with sparse convolutions~\cite{choy20194d}).

\subsection{Multi-task learning}
\begin{figure}[tb]
\begin{center}
    \includegraphics[width=1.0\linewidth,angle=0]{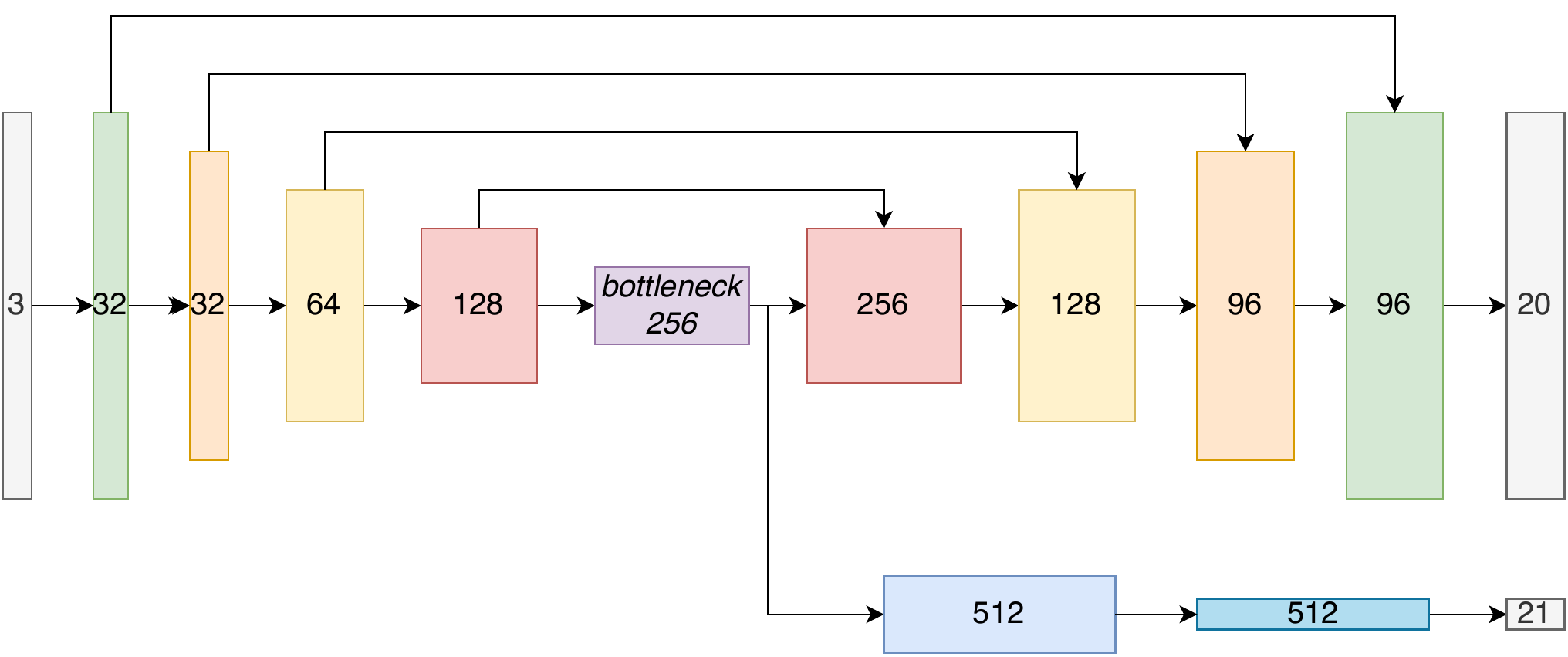}
\end{center}
\caption{The proposed multi-task network architecture has a single encoder, but two output branches, a semantic segmentation head (top) and a scene classification head (bottom).}
\label{fig: network_architecture}
\end{figure}

A limitation for scene recognition is the difficulty of collecting a large enough dataset for deep learning, where "large" should be understood as the number of labels. Since an entire scene -- in the indoor setting often a room -- corresponds to a single example, it is hard to collect many examples per class, and indeed current datasets are limited to $<$100 example per class, on average.
Hence, we resort to multi-task learning. We train our network to, at the same time, also perform per-point semantic labeling of the point cloud, using the same, shared latent representation $Z$.
While that auxiliary task is, arguably, at least as difficult as the scene-level classification, it benefits from a lot stronger supervision with point-wise labels. Therefore, it is reasonable to hope that it will improve the latent feature encoding $Z$, such that the encoding better supports scene classification, too.
For the semantic segmentation we extend the sparse Resnet14 variant of our network with a U-net style decoder that mirrors the encoder, but with a dense set of skip connections (and therefore twice the channel depth of the encoder).
The full architecture is depicted in Fig.~\ref{fig: network_architecture}.

\subsection{Optimisation}
\label{para: optimise}

In our multi-task setting the loss function consists of two terms, a cross-entropy loss $L_{sem}$ for the semantic segmentation and another cross-entropy loss $L_{cls}$ for scene classification. The total loss $L$ is their weighted sum:
\begin{equation}
L=\alpha L_{cls}+(1-\alpha) L_{sem}
\end{equation}
Some care is required to optimise that combined loss. In principle the parameter $\alpha$ balances the two terms. However, due to their vastly different magnitudes the optimisation is very sensitive to the choice of $\alpha$. To see why, consider the huge scale difference -- per one scene label there are thousands of point labels -- and the fact that the loss itself is unaware which points belong to which scene: depending on the value of $\alpha$, flipping a single scene type label can offset many miss-classified points, spread over an entire batch of scenes.
For our task, we found it best to start with the extremes of the weighting, i.e., first train each task separately. First, we set $\alpha=0$ and optimise only for semantic segmentation, where the supervision is much more fine-grained. This will reliably find a reasonable feature encoding in the bottleneck. Then, we freeze the weights of the encoder and train only the scene classification head, effectively treating the semantic segmentation network as a pre-trained feature extractor.
Finally, we fine-tune the network end-to-end, with a small learning rate. Empirically, this schedule turned out to yield the best results.

\section{Experiments}
\subsection{Dataset}
\label{sec: dataset}
\begin{figure*}[tb]
\centering
\begin{minipage}{1\linewidth}
\subfloat[apartment]{\includegraphics[height=1.9cm]{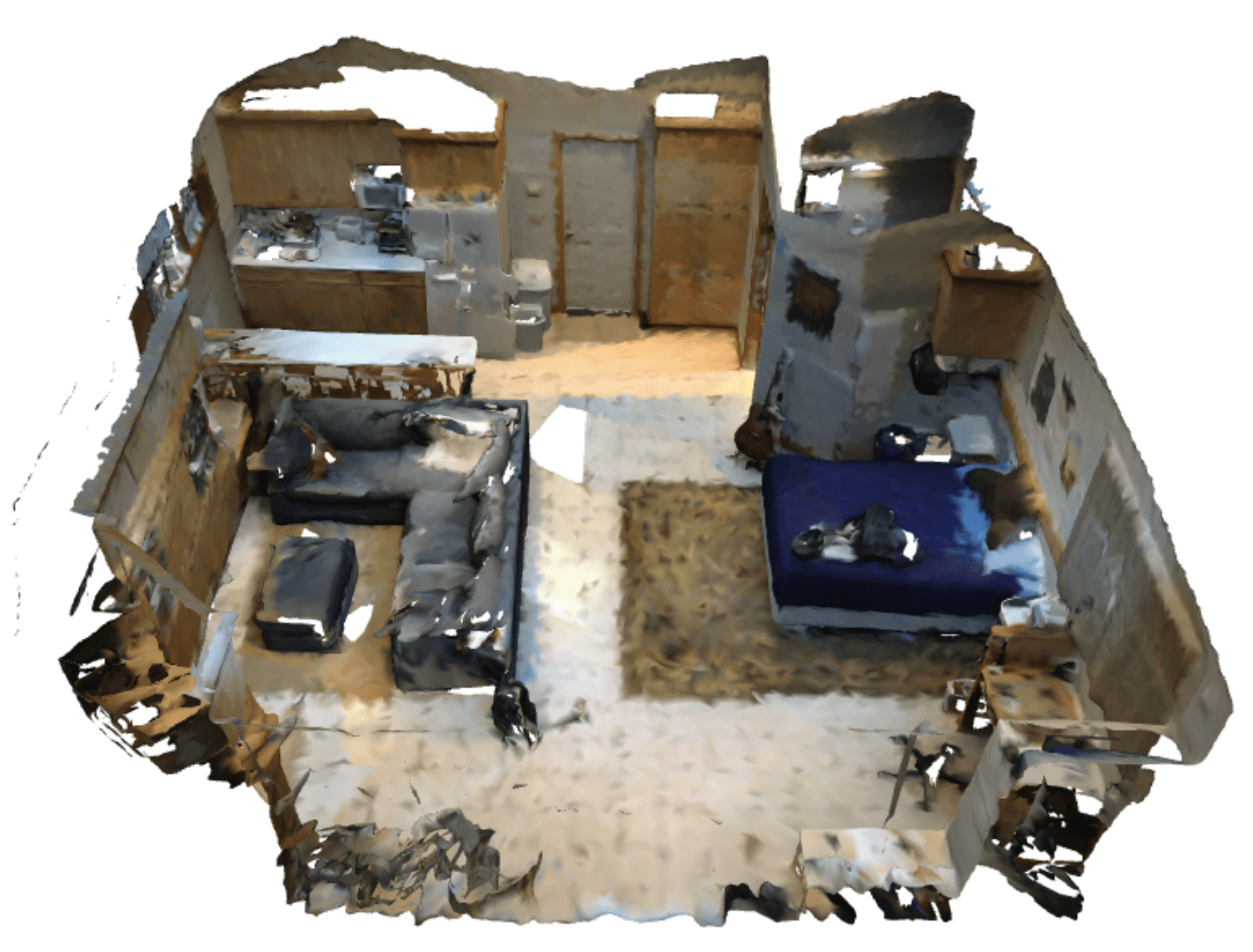}}
\subfloat[bedroom]{\includegraphics[height=1.9cm]{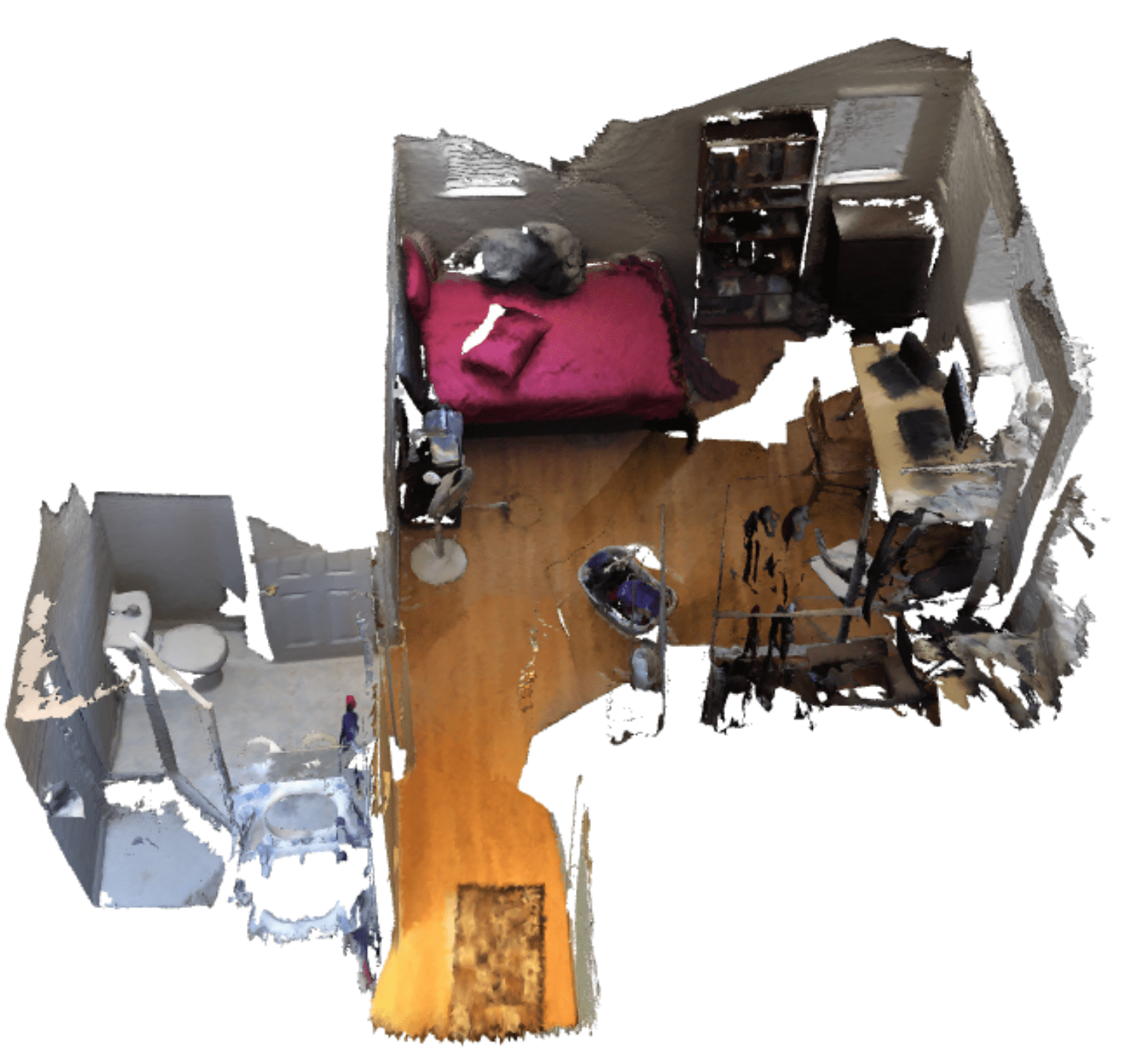}}
\subfloat[living 
room]{\includegraphics[height=1.9cm]{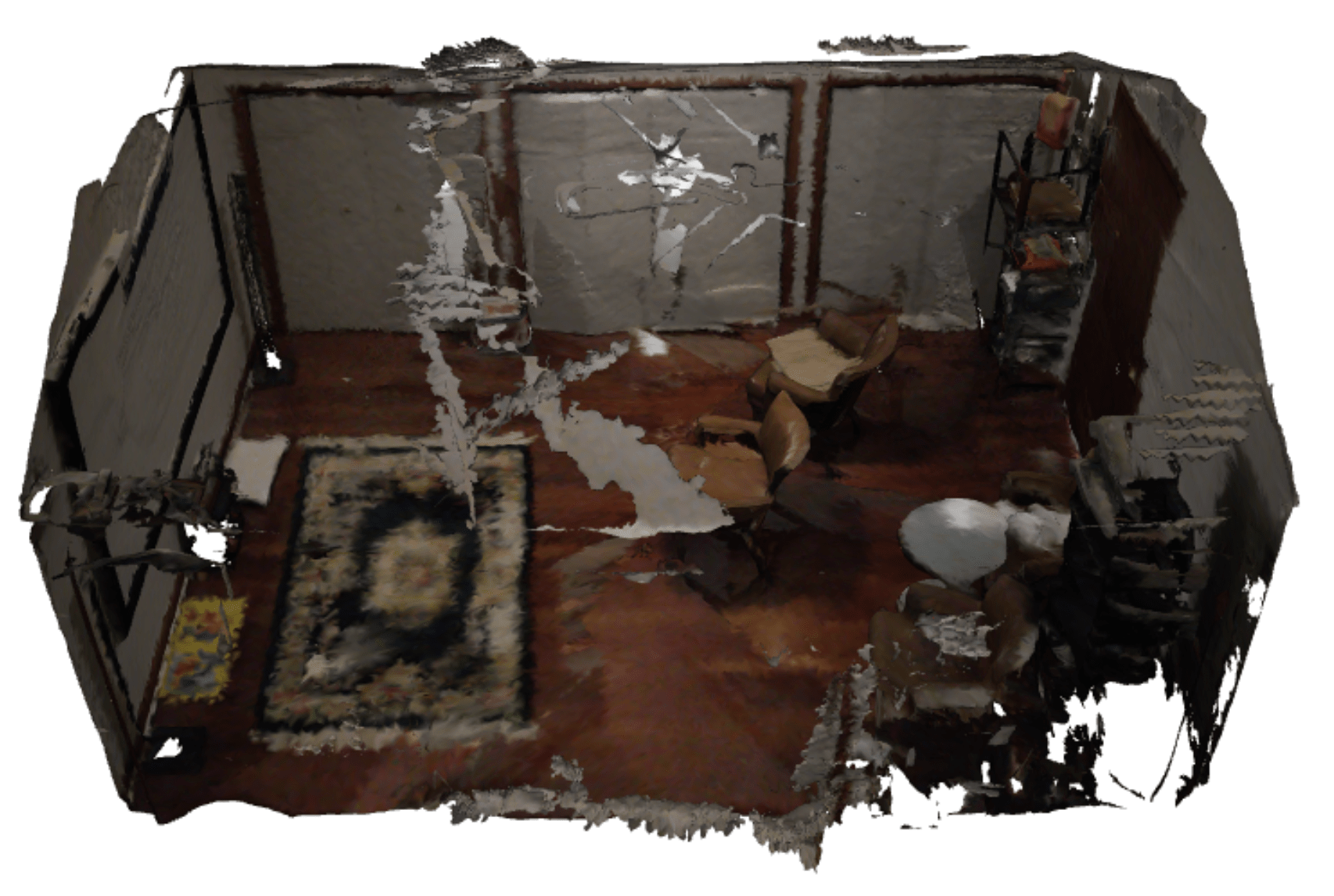}}
\subfloat[kitchen]{\includegraphics[height=1.9cm]{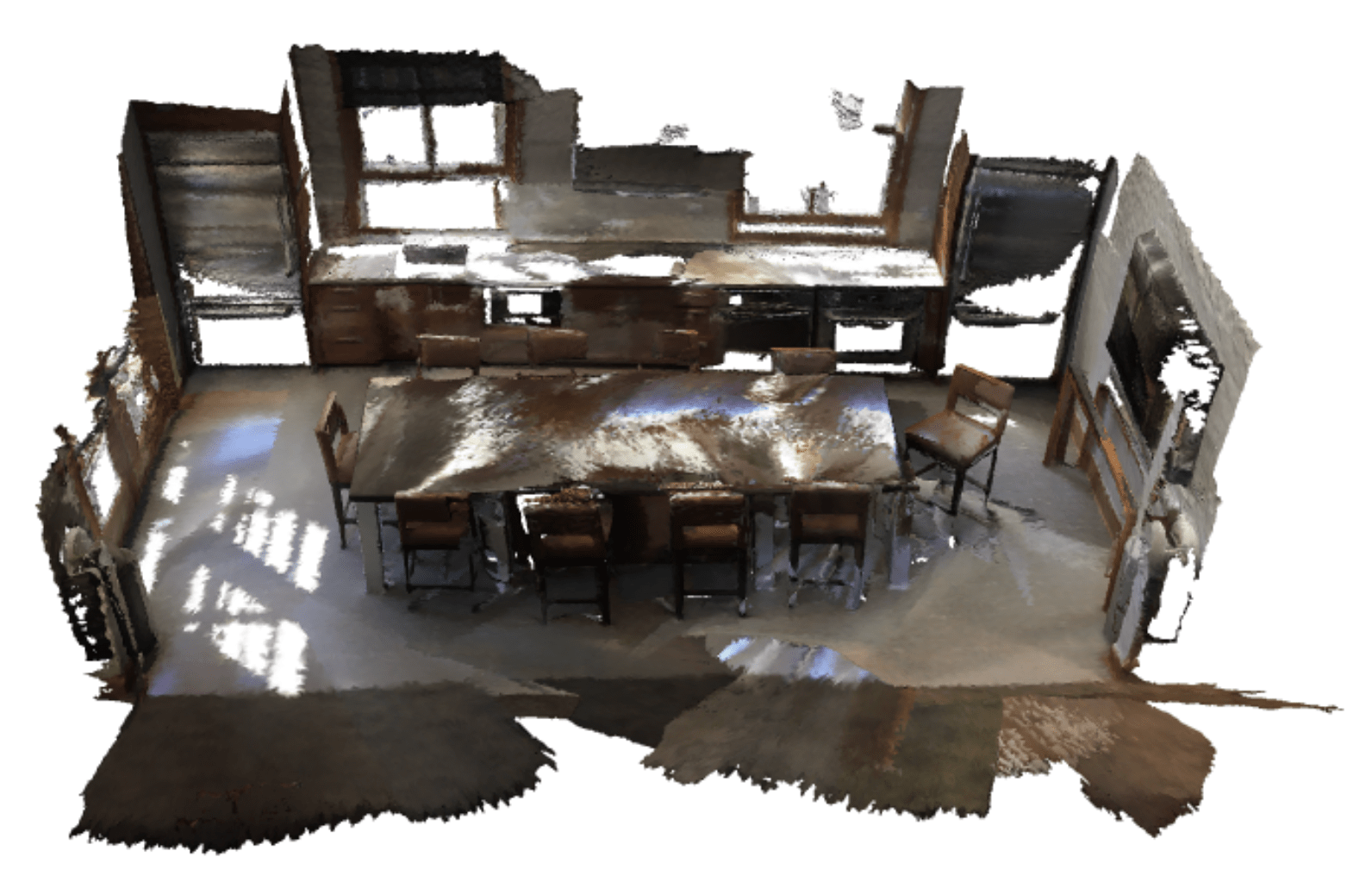}}
\subfloat[conference 
room]{\includegraphics[height=1.9cm]{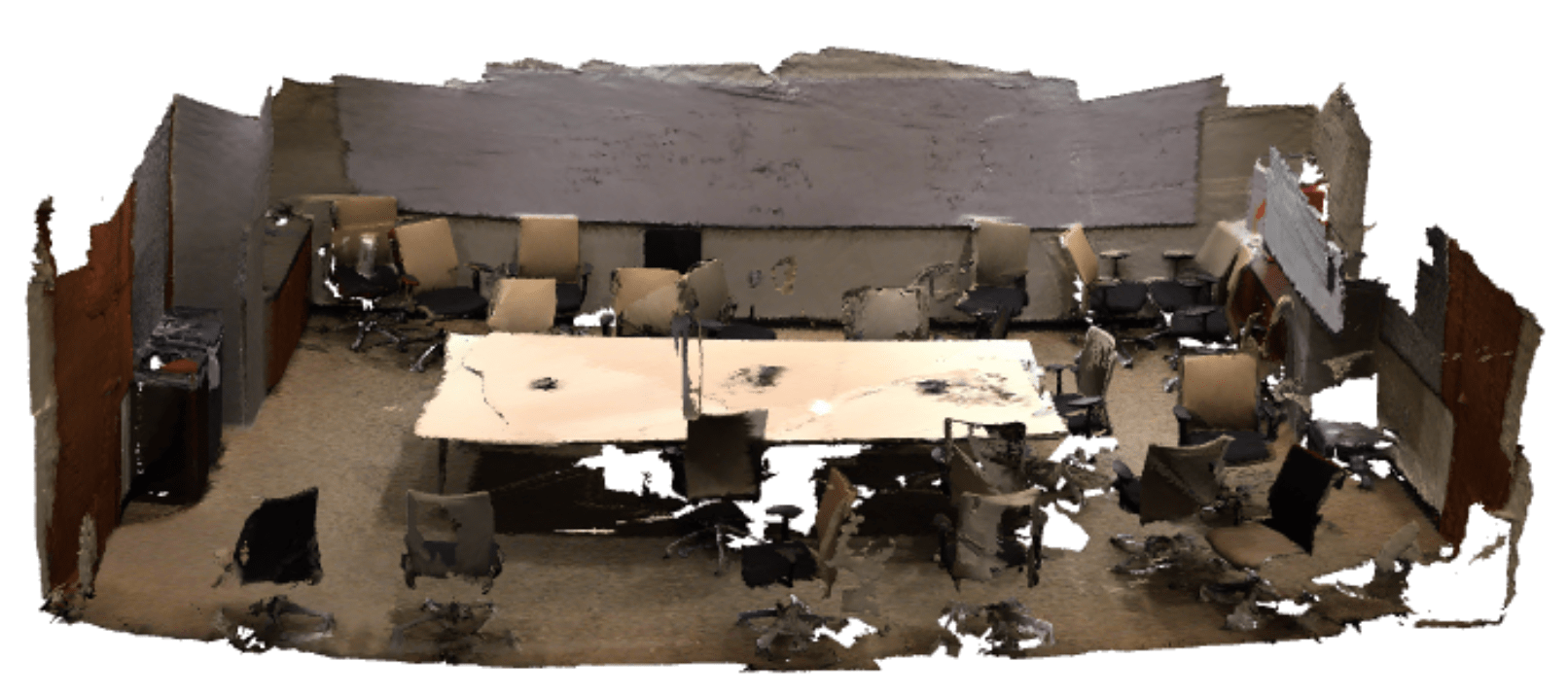}}
\subfloat[copy 
room]{\includegraphics[height=1.9cm]{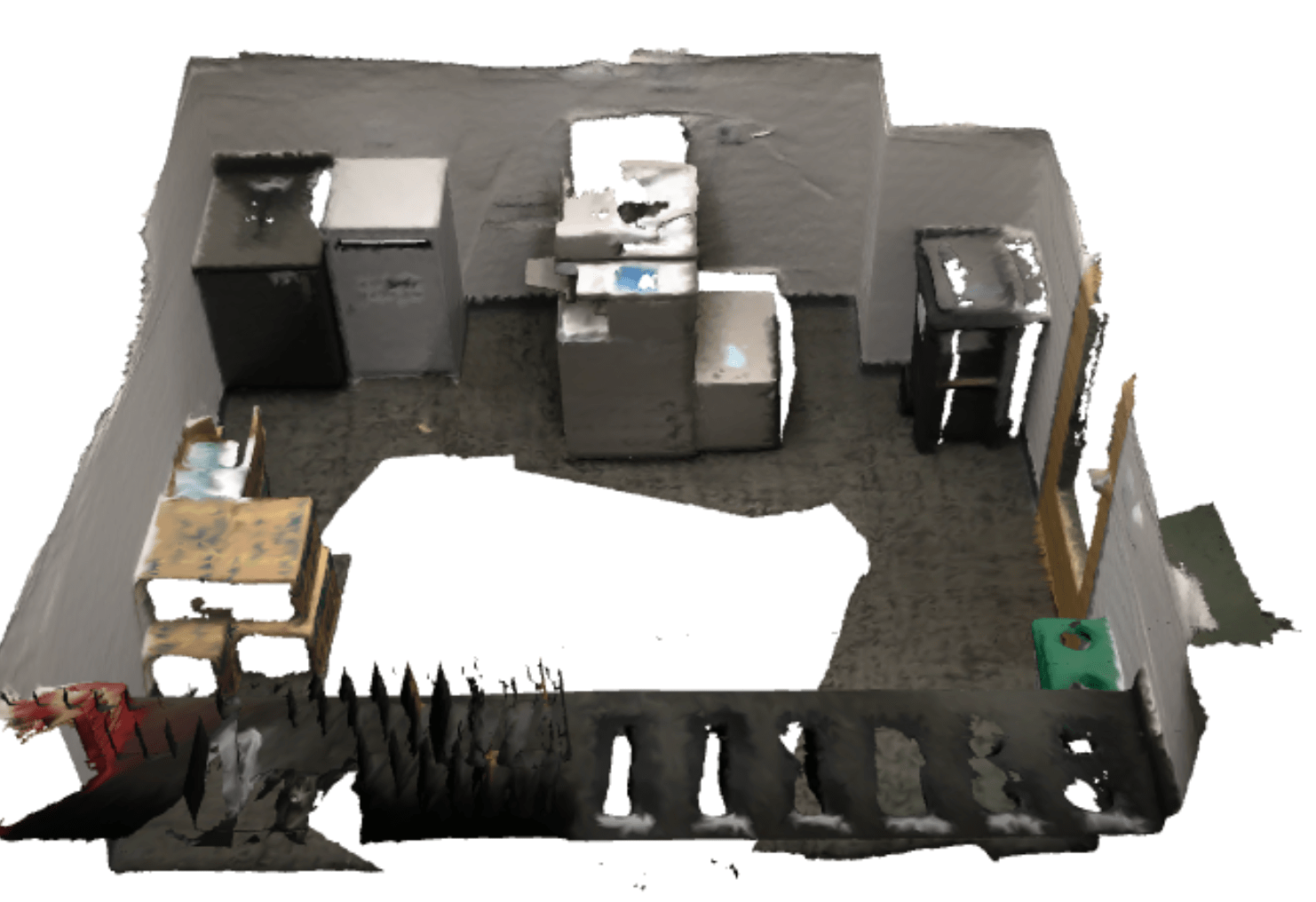}}
\end{minipage}

\begin{minipage}{1\linewidth}
\subfloat[library]{\includegraphics[height=1.9cm]{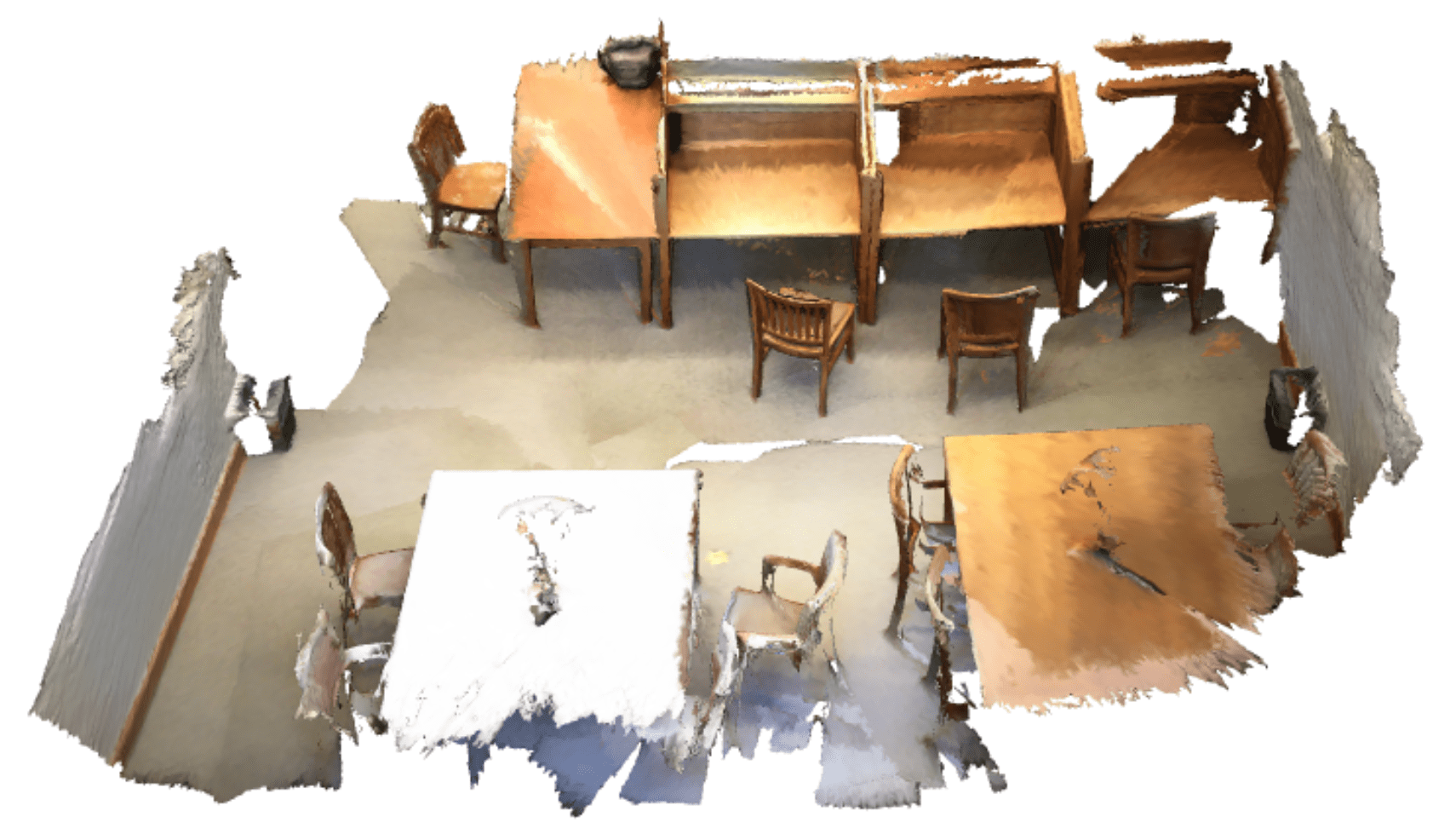}}
\subfloat[office]{\includegraphics[height=1.9cm]{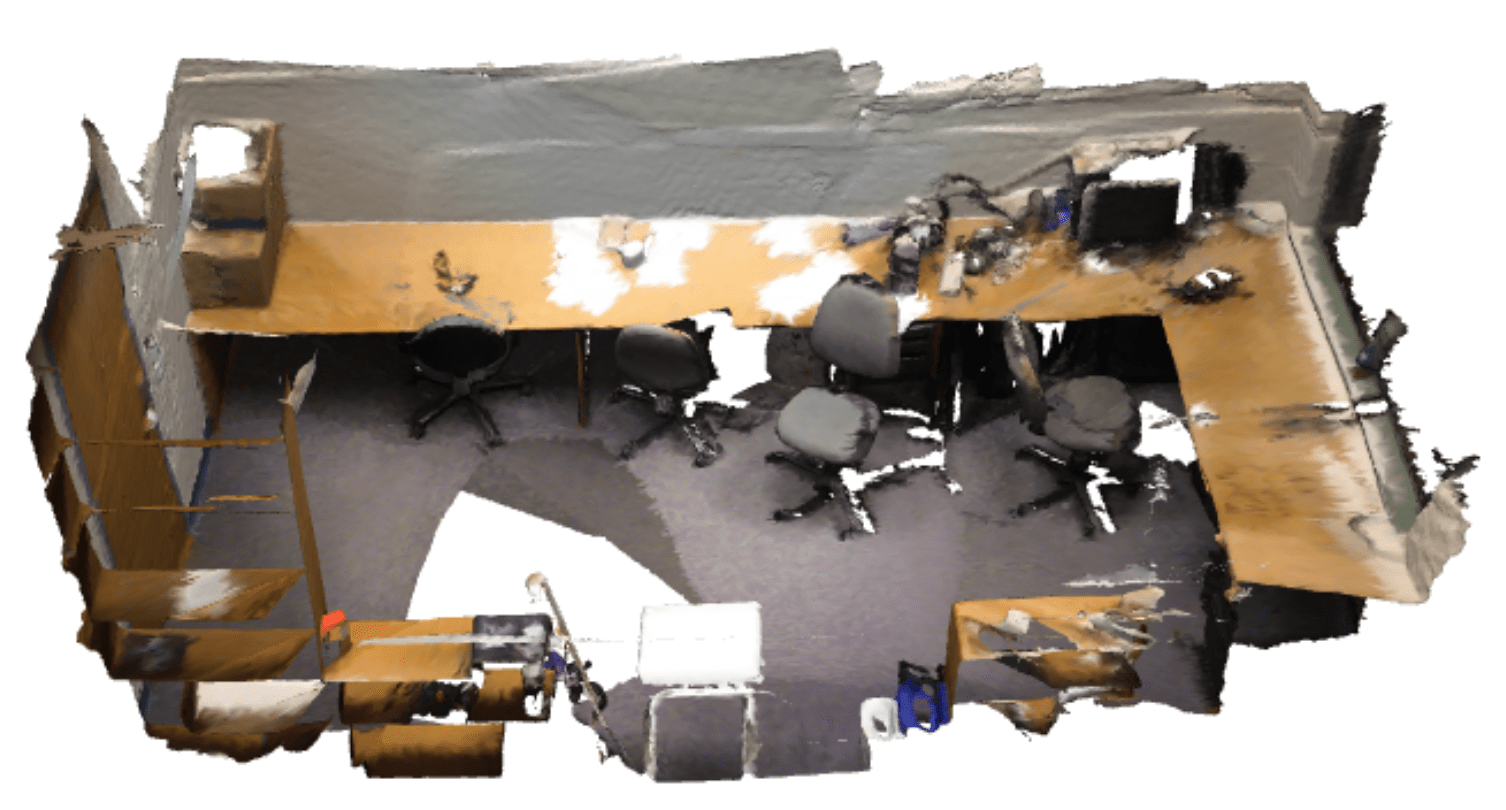}}
\subfloat[hallway]{\includegraphics[height=1.9cm]{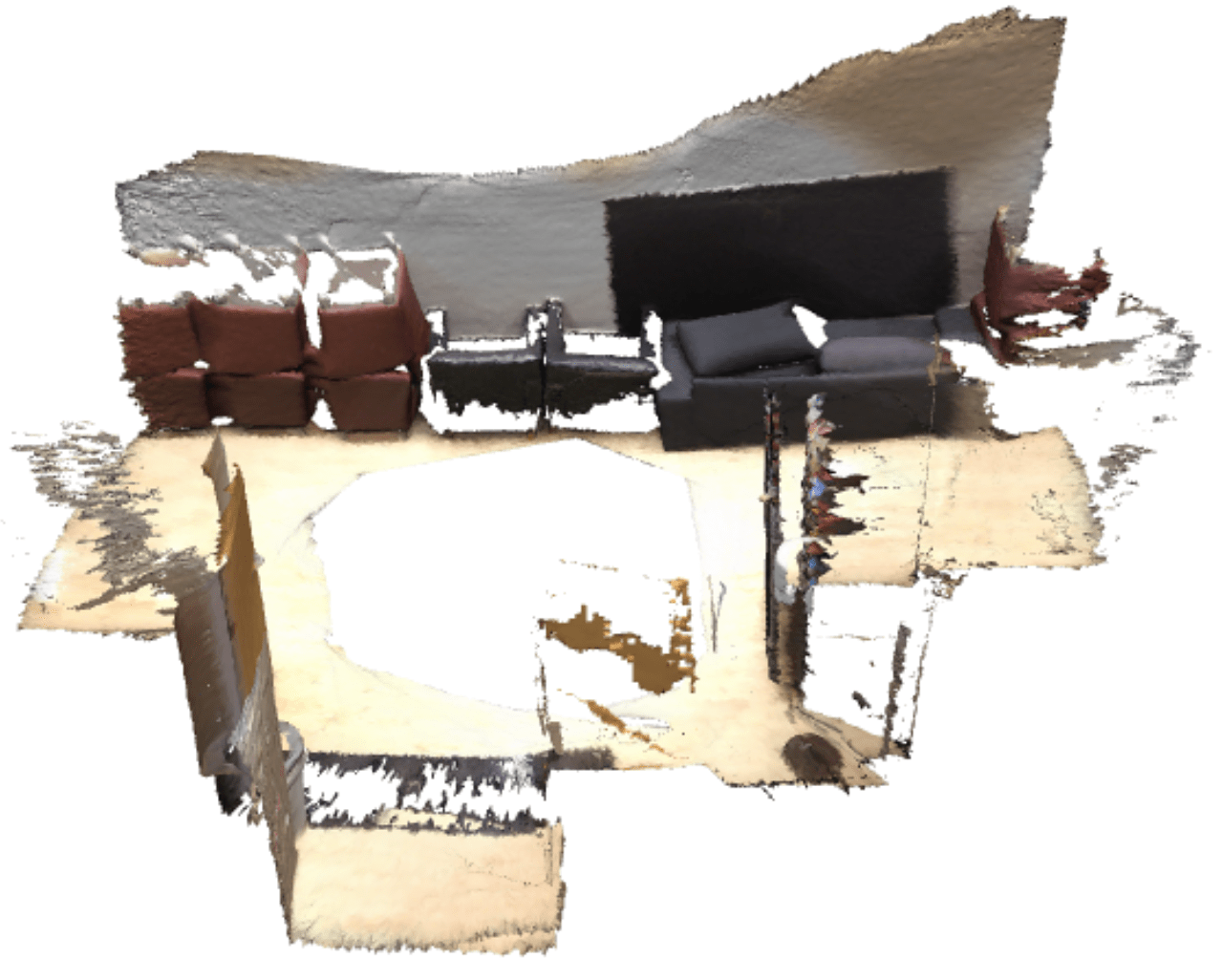}}
\subfloat[storage]{\includegraphics[height=1.9cm]{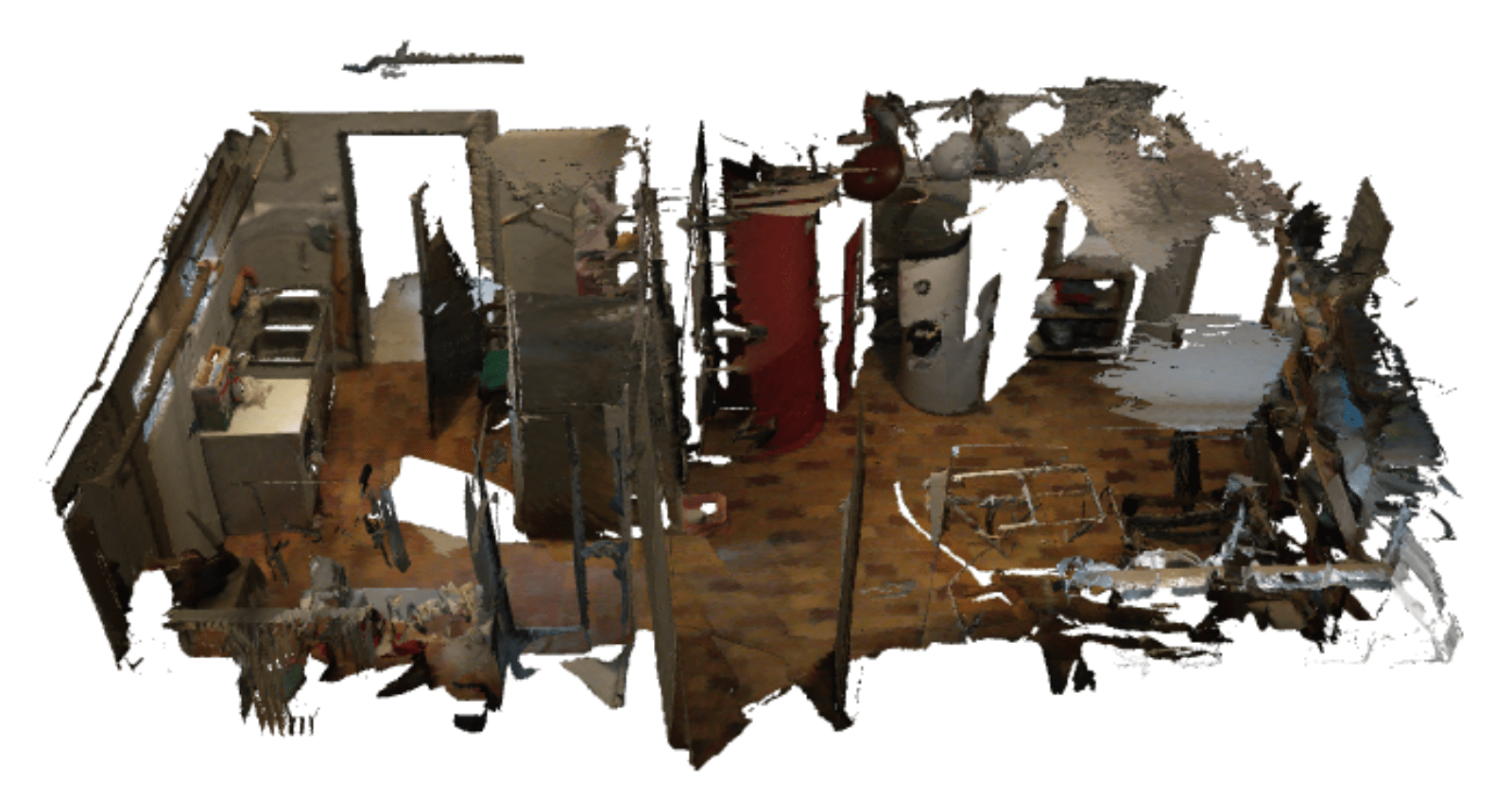}}
\subfloat[bathroom]{\includegraphics[height=1.9cm]{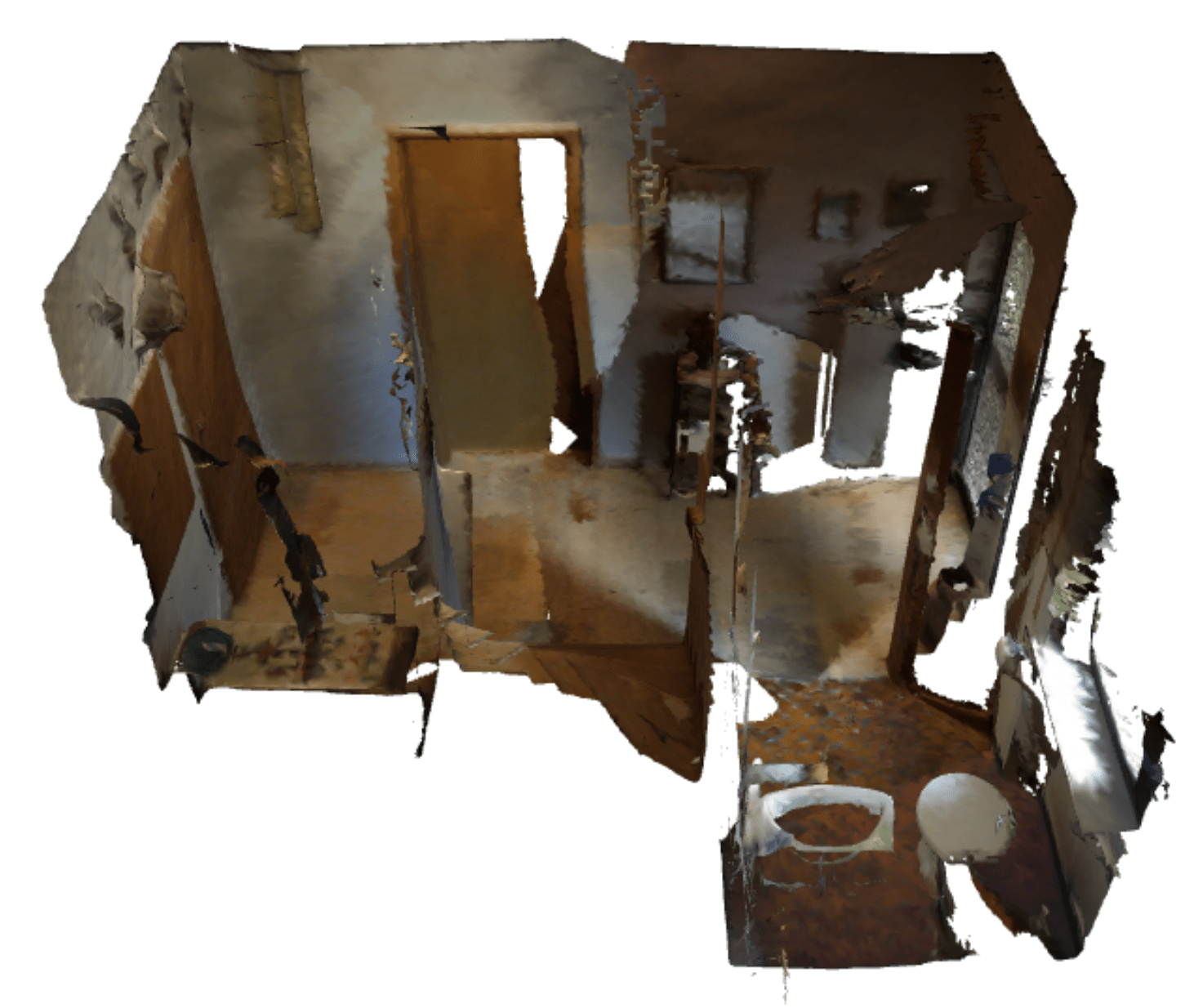}}
\subfloat[laundry]{\includegraphics[height=1.9cm]{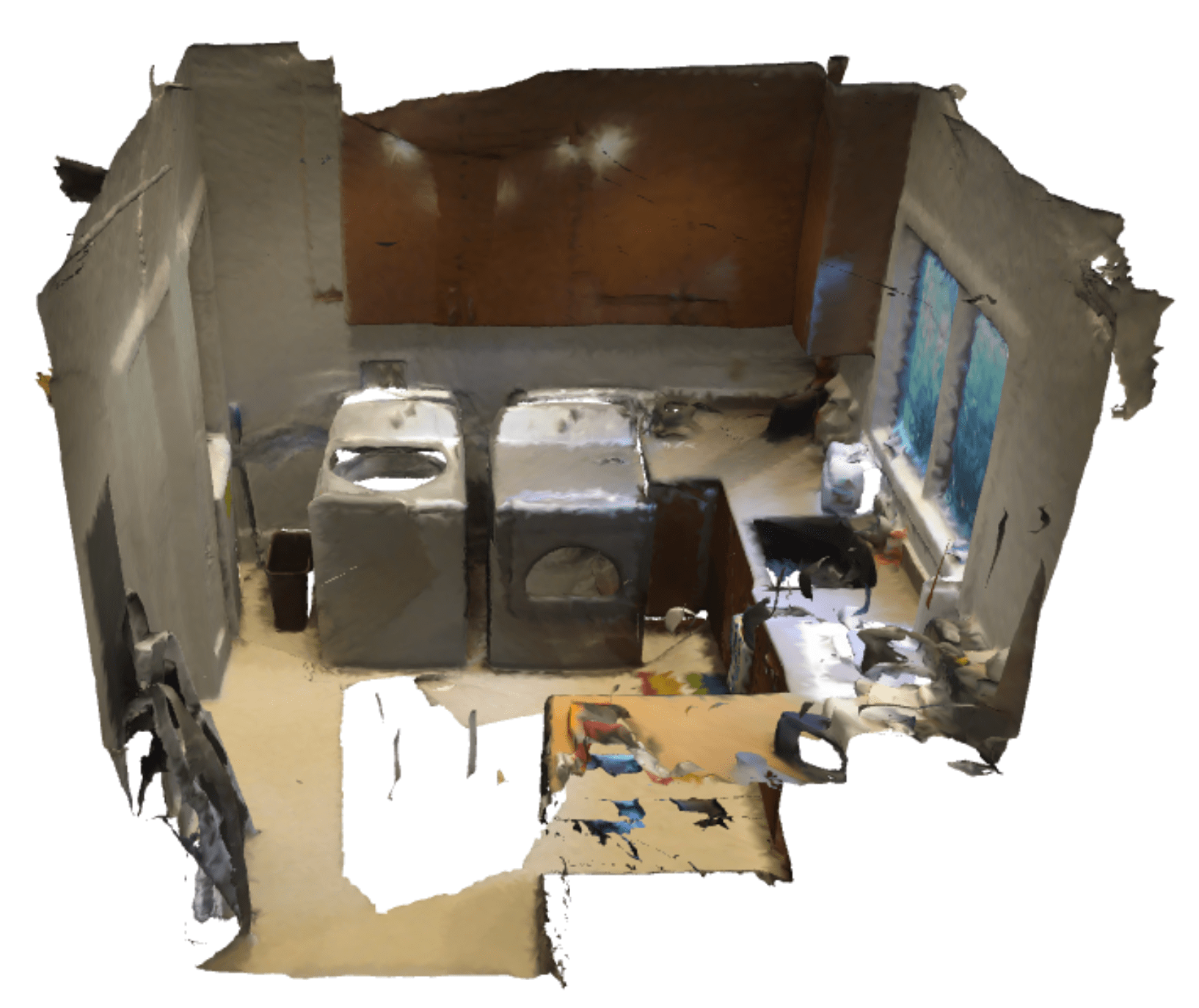}}
\end{minipage}

\caption{Examples of different scene types.}
\label{fig: samples}
\end{figure*}

\begin{figure}[tb]
\begin{center}
  \includegraphics[width=1.0\linewidth,trim=0 25 0 0, clip]{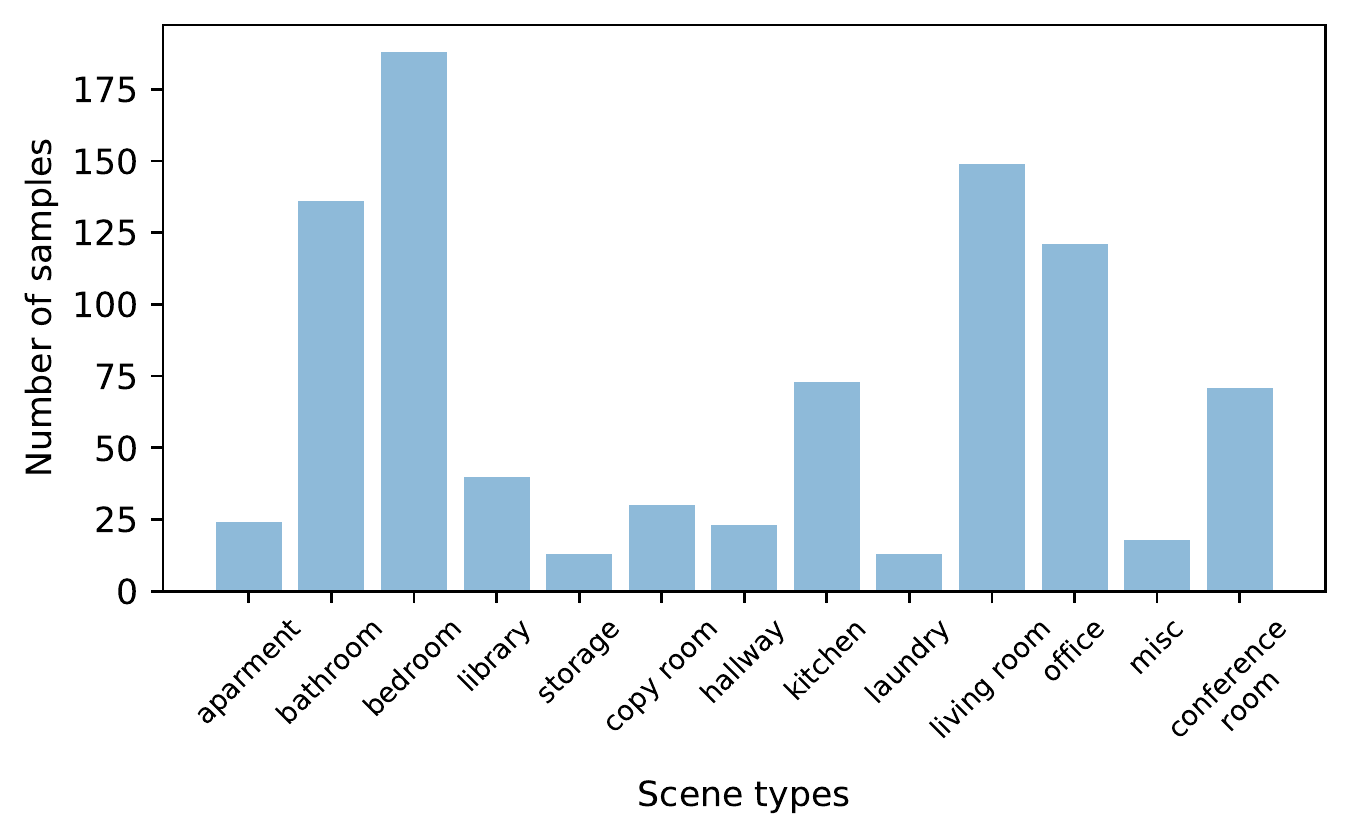}
\end{center}
  \caption{Distribution of the 13 scene types within the training+validation sets.}
\label{fig: dataset}
\end{figure}

ScanNet~\cite{dai2017scannet} is an RGB-D video dataset containing $>$1500 scans. We use 1013 scans for training and 500 scans for validation. A further 100 scans are designated as test set for the benchmark, for these the ground truth is withheld. Fig.~\ref{fig: samples} shows examples for different scene types. On average, one scan has 150k points and a spatial extent of 5.5m $\times$ 5.1m $\times$ 2.4m. For the scene classification task, there are in total 21 classes, of which only a subset of 13 classes is evaluated in the benchmark (but we always predict all 21 logits). Fig.~\ref{fig: dataset} shows the class distribution for the reduced 13-class set. As it is quite imbalanced, we report both accuracy (fraction of correct classifications) and mean Intersection-of-Union (i.e., compute IoU values per scene type and average them).

\subsection{Baselines}
As sanity check for the different deep learning models, we construct three baselines that completely ignore the spatial layout. First, we represent each scene by its 30-bin colour histogram and label the test scans with nearest-neighbour classification (which worked slightly better than a Random Forest classifier). This yields a classification accuracy of 57.1\%, a fairly high success rate for such a naive scheme, which we regard as a lower bound for any more sophisticated approach.
Second, we predict semantic labels with the U-net described above (without scene classification head), turn the maximum-likelihood labels into a normalised object class histogram and train a Random Forest classifier on those histograms. This achieves a strong 82.8\% accuracy, indicating that the statistics over point labels without any spatial information encodes the scene type rather well.
Finally, as an upper bound for these "geometry-free" approaches we pretend to have a oracle for semantic segmentation and train the same Random Forest classifier on the ground truth label histograms. This yields 85.0\% accuracy, meaning that the errors of the semantic segmentation only slightly degrade the performance.

% \begin{table}[H]
% \centering
% \resizebox{1\linewidth}{!}{
% \begin{tabular}{|c|c|c|c|c|}
% \hline
%                     & KNN(K=1)(\%) & Linear SVM(\%) & Random Forest(\%) & MLP(\%) \\ \hline
% semantics histogram & 78.8         & 57.1           & 84.0              & \textbf{84.6}    \\ \hline
% instance histogram  & 79.1         & 56.7           & 83.8              & 82.1    \\ \hline
% \end{tabular}}
% \caption{\textbf{Hard baseline.} We use all the default parameters in Sklearn~\cite{scikit-learn}, and MLP has only 1 hidden layer with 128 neurons. The metric is accuracy.}
% \label{tab: baseline}
% \end{table}

\subsection{Deep learning models}
\label{sec: deep learning}

% \subsubsection{Iterative Farthest Point Sampling} 
% \label{sec: FPS}
% Sub-sampling is important for Pointnets and DGCNN~\cite{wang2019dynamic} which require fixed number of points as input. Given input points $\{x_1, x_2, ..., x_n\}$, we use iterative farthest point sampling (FPS)~\cite{qi2017pointnet++} to choose a subset of points $\{x_{i_1}, x_{i_2}, ..., x_{i_m}\}$, such that $x_{i_j}$ is the most distant point (in metric distance) from the set $\{x_{i_1}, x_{i_2}, ..., x_{i_{j − 1}}\}$ with regard to the rest points. Compared with random sampling, it has better coverage of the entire point set given the same number of centroids. We take the implementation from this repository. For vanile PointNet, PointNet++ and DGCNN, we use FPS to sample 4096 points from raw point cloud. 

% \subsubsection{Voxelisation} 
% \label{sec: voxelisation}
% Voxelisation is a key step in voxel-based methods. Given a point cloud, by defining the voxel resolution, we can discretize points live in continuous space to discrete space. When the voxel resolution is big, it's common that several points with different labels/colors fall into the same voxel. MinkowskiEngine randomly chooses one point and assigns its label to the voxel by default. SparseConv supports random picking one label or averaging the labels. We set the voxel resolution to be 10cm, which would bring roughly the same number of non-empty voxels as 4096.

\subsubsection{Implementation details} 
\label{sec: imp_details}
Pointnets and DGCNN require a fixed number of points as input. Unless stated otherwise, we sample 4096 points per scan using Farthest Point Sampling for optimal coverage.
For the sparse, voxel-based Resnet14 we fix the voxel size to 2cm. When points with different labels/colours fall into the same voxel, we randomly pick one.
All models are trained with batch size 16 using the Adam~\cite{kingma2014adam} optimiser with base learning rate $10^{-3}$. For  vanilla Pointnet and DGCNN we use the cosine annealing learning rate scheduler. All models are trained for 300 epochs on a single GeForce GTX 1080Ti except DGCNN, for which we use 3 GPUs, as it is computationally expensive.

\subsubsection{Data augmentation} 
\label{sec: aug_data}
As mentioned, the number of scenes is limited, for several scene types there are $<$25 training exemplars. Therefore, we use multiple forms of data augmentation. We randomly translate the point cloud, rotate it around the vertical axis, scale it by random factors between 0.8 and 1.25, add Gaussian noise to the point coordinates, and randomly remove 12.5\% of the points.
Finally, we also randomly cut away points in one corner of the bounding box, inspired by cutout regularisation~\cite{devries2017improved} for images. 
To ensure reproducibility, we publish all our code.%
\footnote{\url{https://github.com/ShengyuH/Scene-Recognition-in-3D}}

\subsection{Results}
\label{sec: results}
The overall results on our validation set are displayed in Table \ref{tab: comparison}. 
Pointnet in its basic form and DGCNN perform worst, in particular Pointnet with only geometry and no colour information performs on par with the simple colour histogram baseline, while adding colour is particularly beneficial for Pointnet.
Variants of Pointnet++ and  sparse Resnet14 achieve much better results between 83.6\% and 87.8\%. Interestingly, Pointnet++ performs better without colour, while Resnet14 performs better with colour information.
Note, the big gap between Pointnet and Pointnet++ indicates that locality and spatial layout play a role for scene recognition, as the global pooling of Pointnet is detrimental.

With 90.3\% accuracy, our proposed multi-task learner achieves the best performance among these models. Its results can be directly compared to Resnet14, which has the same encoder and scene classification head, without the auxiliary semantic segmentation head.
We have also submitted results of our best network to the ScanNet benchmark. The numbers differ significantly from those on our validation set, presumably due to the small size of the withheld test set (100 scans spread across 13 classes). 
Detailed per-class results are shown in Table \ref{tab: detailed_results}. 
The only two prior submissions were based on 2D (birds-eye) views and achieved an average recall of at most 49.8\%, respectively a mean IoU of 35.5\%. We outperform them by a large margin: with our sparse Resnet14 that solves the task directly in 3D, supported by multi-task learning, performance jumps to 70.0\% recall, respectively 64.6\% mean IoU.

\begin{table}[tb]
\caption{Results of deep learning models.}
\label{tab: comparison}
\resizebox{\columnwidth}{!}{
\begin{tabular}{ll|c c c}
\hline
\multicolumn{2}{c}{}                                 & Acc [\%] & mIoU [\%]       & params [M] \\ \hline
\multicolumn{2}{l|}{Colour histogram} & 57.1 & 36.1 & - \\ \cline{1-2}
\multicolumn{2}{l|}{Point class histogram (pred)} & 82.8 & 70.2 & - \\
\multicolumn{2}{l|}{Point class histogram (oracle)} & 85.0 & 73.9 & - \\ \cline{1-2}
       Pointnet         & XYZ only    & 57.4          & 35.6          & 0.68   \\ 
                         & XYZ+colour & 70.8          & 52.2          & 0.68   \\ \cline{1-2} 
DGCNN                   & XYZ only    & 77.2          & 54.5          & 1.80\\ 
                        & XYZ+colour & 77.0          & 62.1          & 1.81         \\ \cline{1-2} 
Pointnet++              & XYZ only    & 87.8          & 75.1          & 1.74\\  
                        & XYZ+colour & 84.8          & 69.1          & 1.74\\ \cline{1-2} 
Resnet14                & XYZ only    & 83.6          & 67.6          & 21.4\\  
                        & XYZ+colour & 87.4          & 70.6          & 21.5\\ \cline{1-2} 
Multi-task            & XYZ+colour & \textbf{90.3} & \textbf{77.2} & 35.3         \\ \hline
\end{tabular}}
\end{table}

\begin{table*}[tb]
\caption{Detailed scene classification results on ScanNet dataset. Numbers 
reproduced from the evaluation server 
\url{http://kaldir.vc.in.tum.de/scannet_benchmark/}}
\label{tab: detailed_results}
\resizebox{\textwidth}{!}{
\begin{tabular}{l|ccccccccccccc|c}
\hline
Method            & apartment      & bathroom       & bedroom & library        & \begin{tabular}[c]{@{}c@{}}conference \\ room\end{tabular} & \begin{tabular}[c]{@{}c@{}}copy \\ room\end{tabular} & hallway        & kitchen        & \begin{tabular}[c]{@{}c@{}}laundry\\ room\end{tabular} & \begin{tabular}[c]{@{}c@{}}living \\ room\end{tabular} & misc  & office         & storage        & \begin{tabular}[c]{@{}c@{}}avg\\ recall\end{tabular} \\ \hline
resnet50\_scannet~\cite{dai2017scannet} & 0.250       & 0.812       & 0.529       & {\bf 0.500} & 0.500       & 0.000       & {\bf 0.500} & 0.571       & 0.000       & 0.556      & 0.000        & 0.375       & {\bf 0.000} & 0.353\\ %\cline{1-1}
SE-ResNeXT-SSMA~\cite{valada2019self}   & 0.000       & 0.812       & {\bf 0.941} & {\bf 0.500} & 0.500       & 0.500       & {\bf 0.500} & 0.429       & 0.500       & 0.667       & {\bf 0.500} & 0.625       & {\bf 0.000} & 0.498\\ %\cline{1-1}
Ours                                    & {\bf 0.500} & {\bf 1.000} & 0.882       & {\bf 0.500} & {\bf 1.000} & {\bf 1.000} & {\bf 0.500} & {\bf 1.000} & {\bf 1.000} & {\bf 0.778} & 0.000       & {\bf 0.938} & {\bf 0.000} & {\bf 0.700}\\ \hline
\end{tabular}}
\end{table*}

\section{Analysis and discussion}
We run a number of further experiments to analyse the 3D approach in more detail. 

\subsection{Sparsity}
\begin{figure}[tb]
\begin{minipage}{1\linewidth}
\centering
\subfloat[32]{\includegraphics[height=1.8cm]{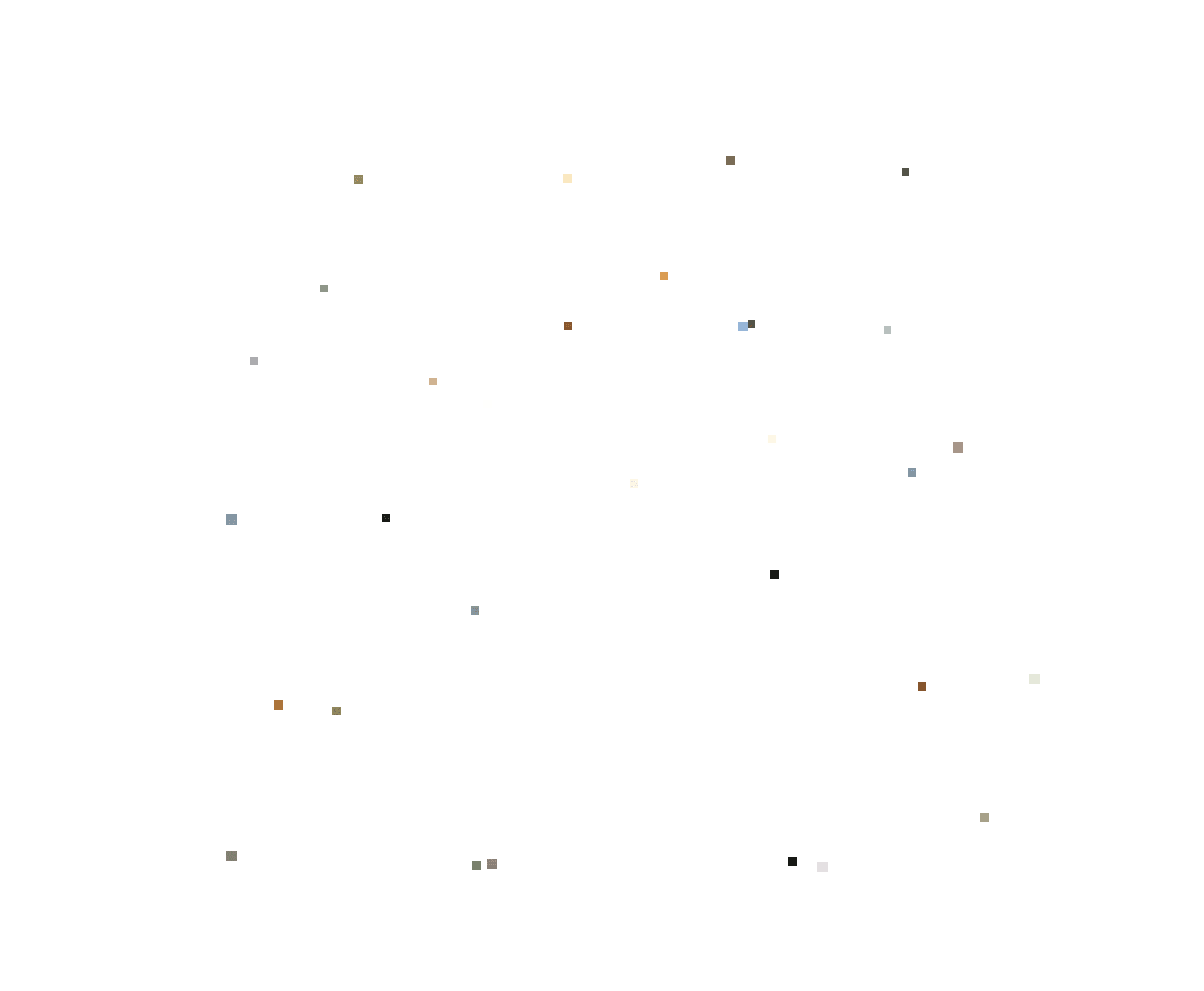}}
\subfloat[128]{\includegraphics[height=1.8cm]{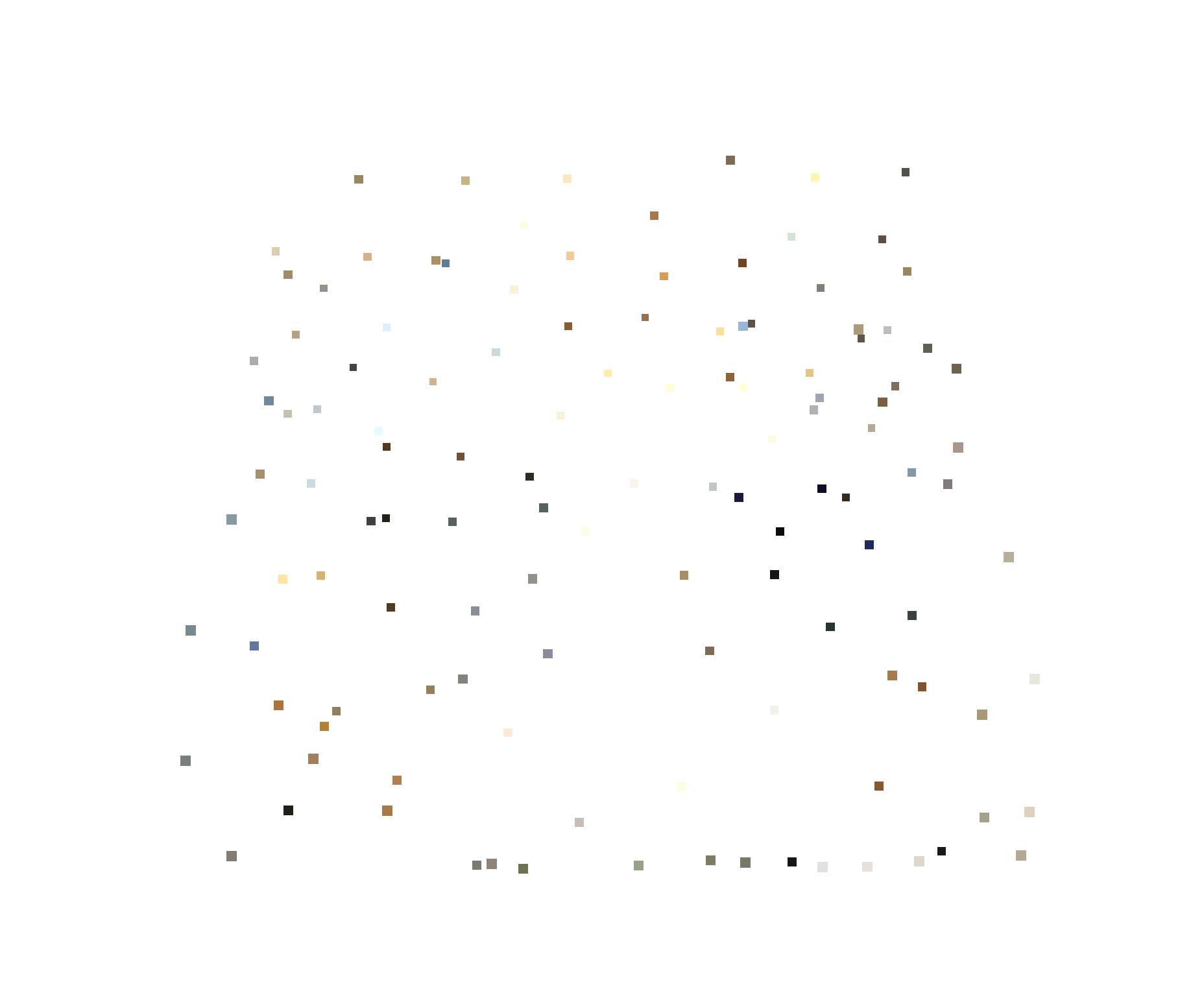}}
\subfloat[512]{\includegraphics[height=1.8cm]{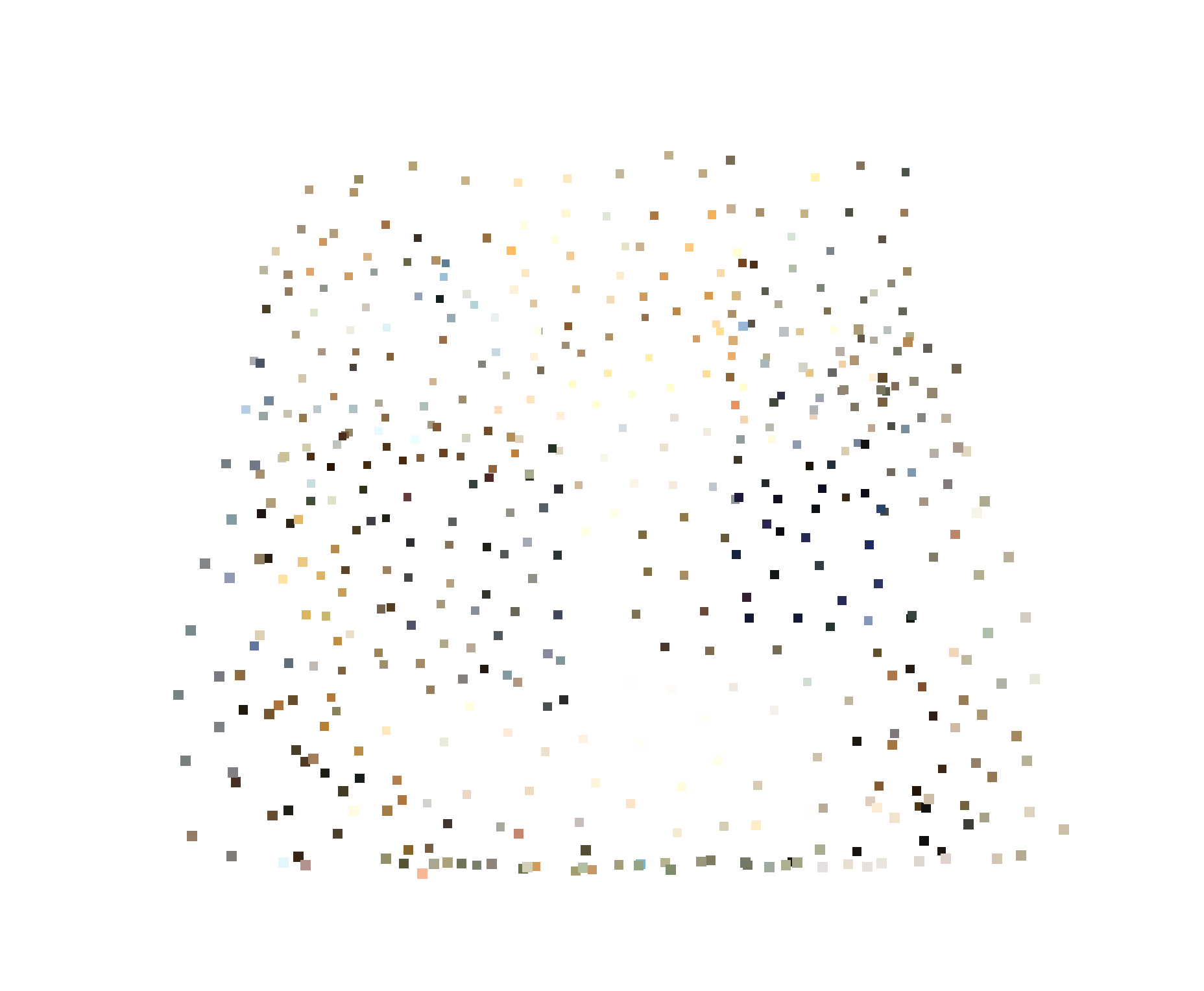}}
\subfloat[1024]{\includegraphics[height=1.8cm]{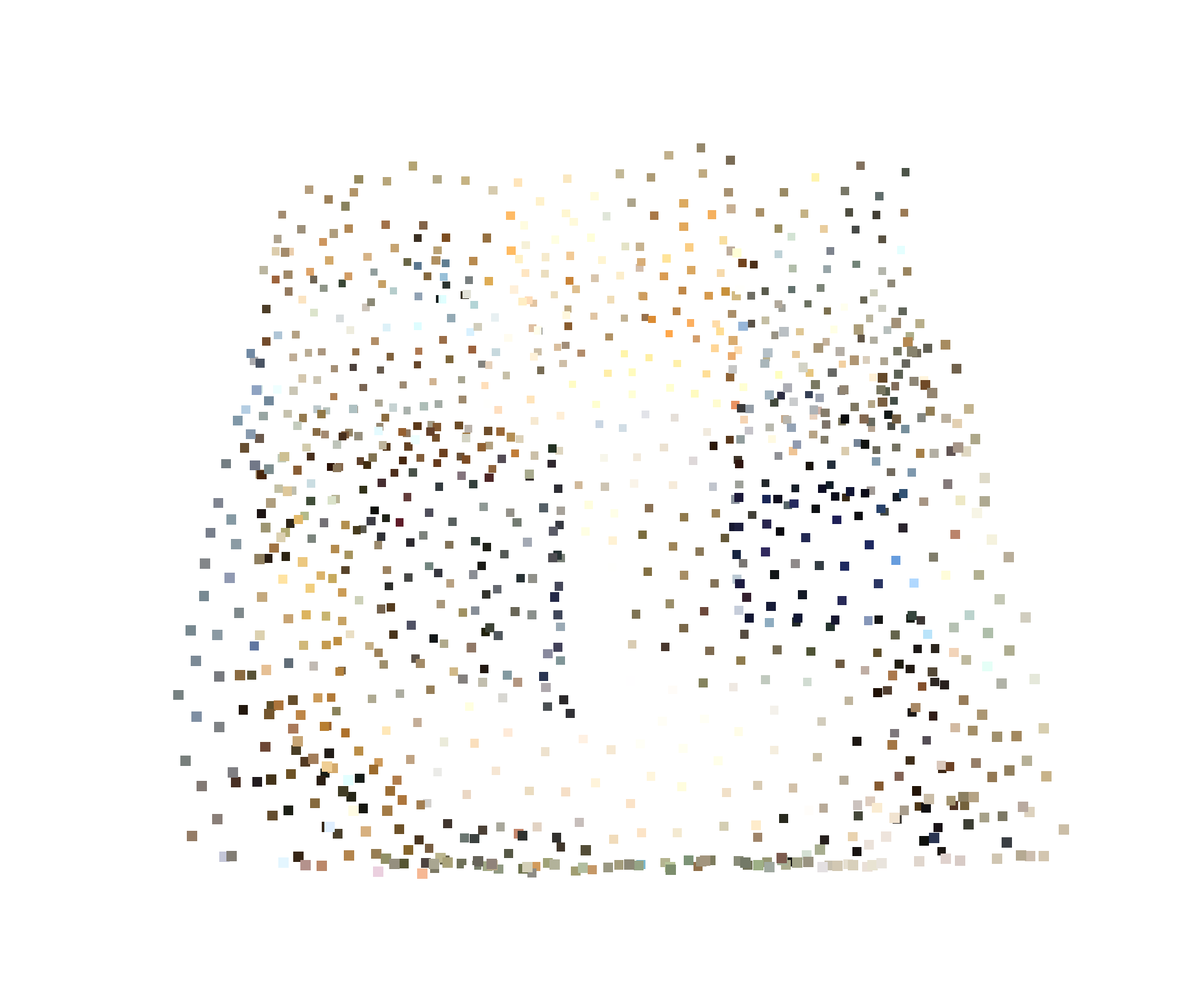}}
\end{minipage}
\begin{minipage}{1\linewidth}
\centering
\subfloat[2048]{\includegraphics[height=1.8cm]{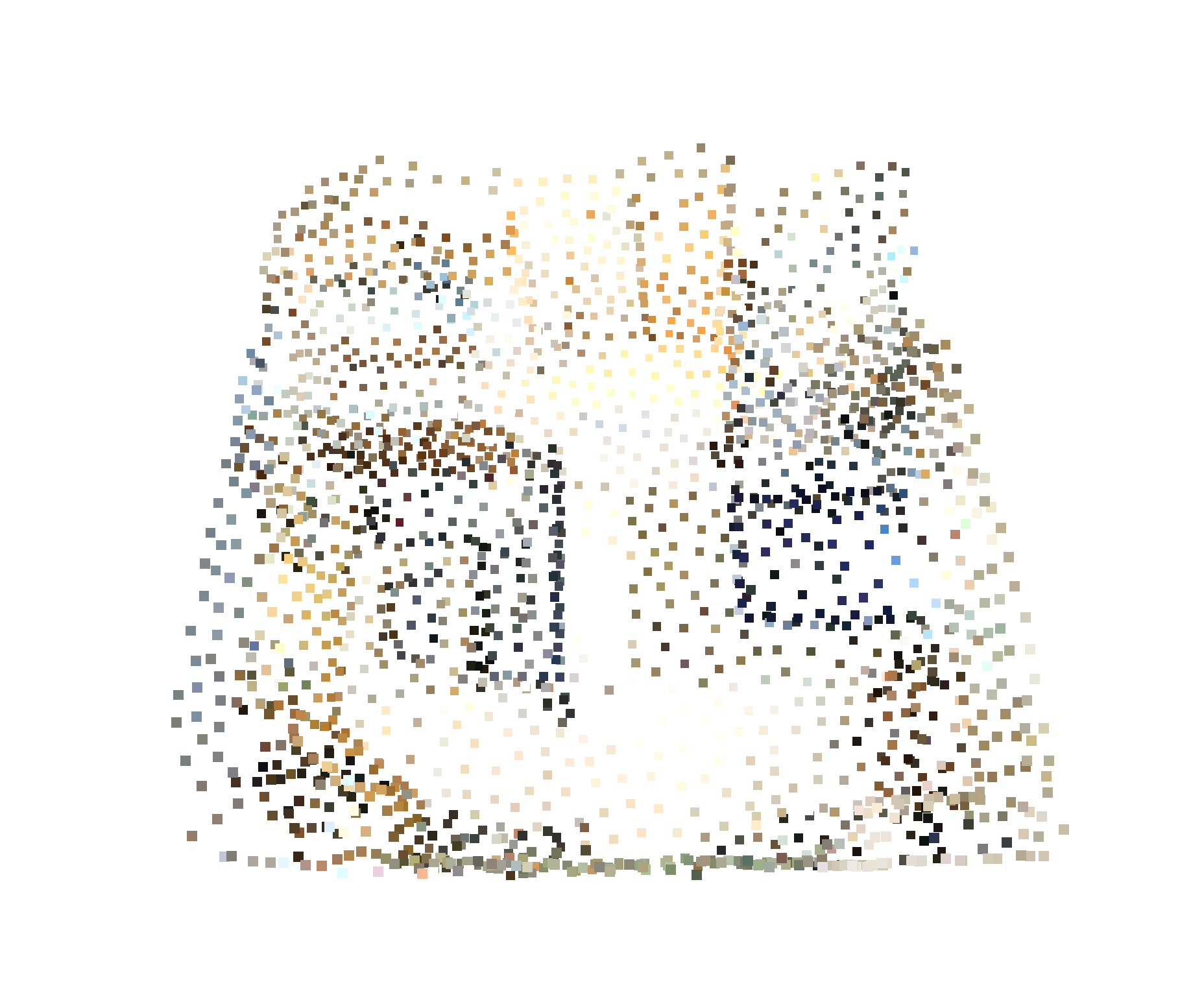}}
\subfloat[4096]{\includegraphics[height=1.8cm]{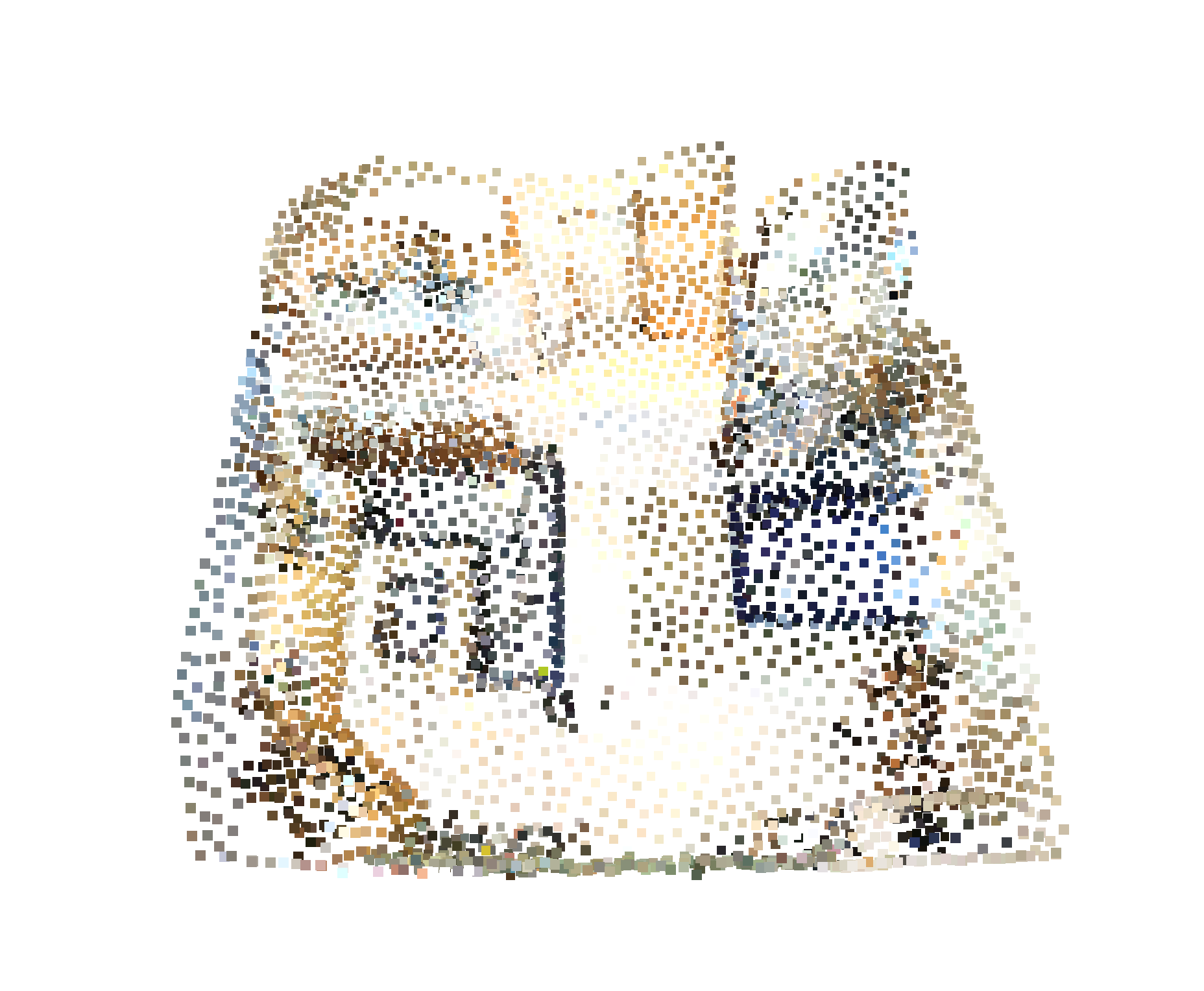}}
\subfloat[16184]{\includegraphics[height=1.8cm]{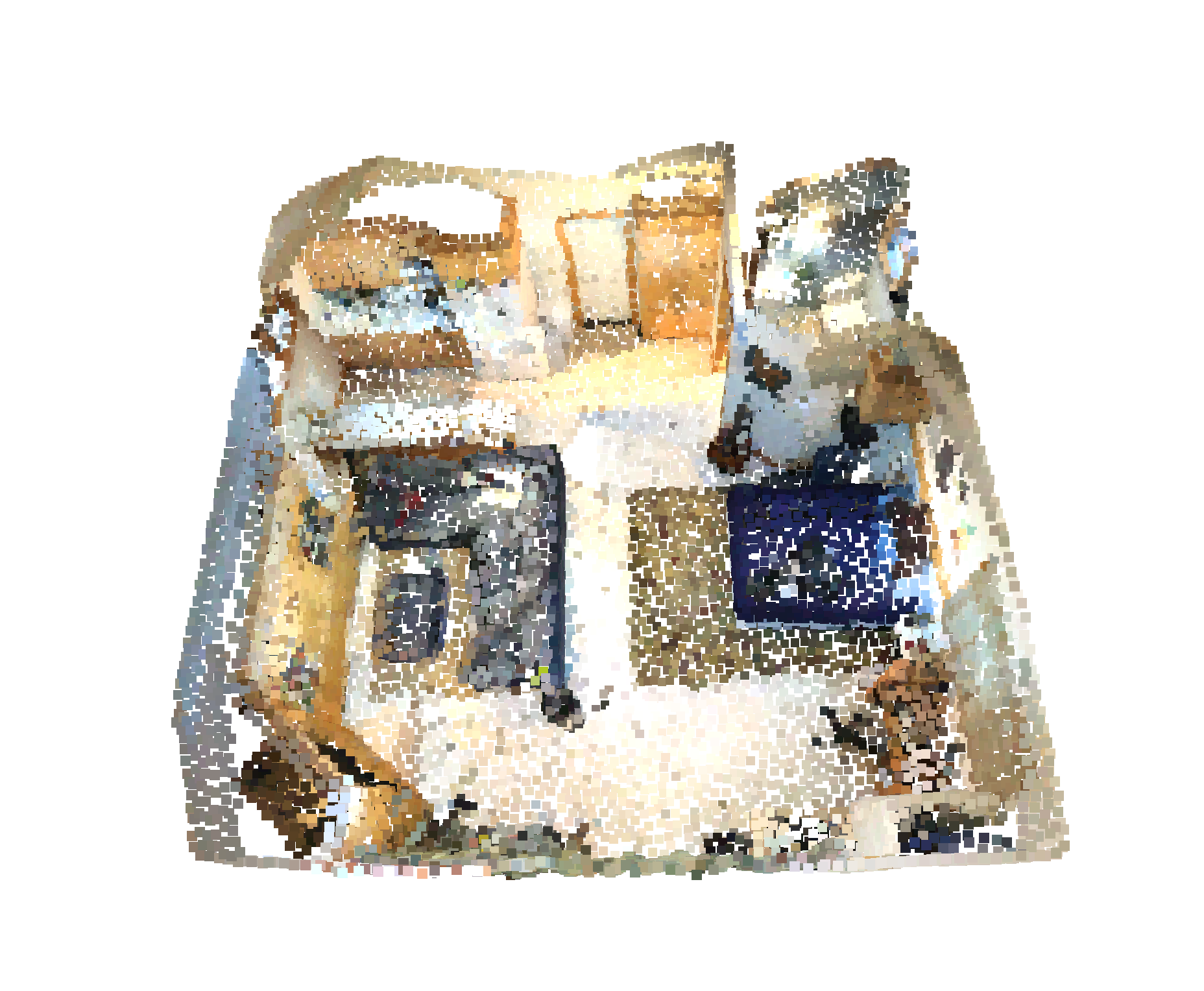}}
\subfloat[81369]{\includegraphics[height=1.8cm]{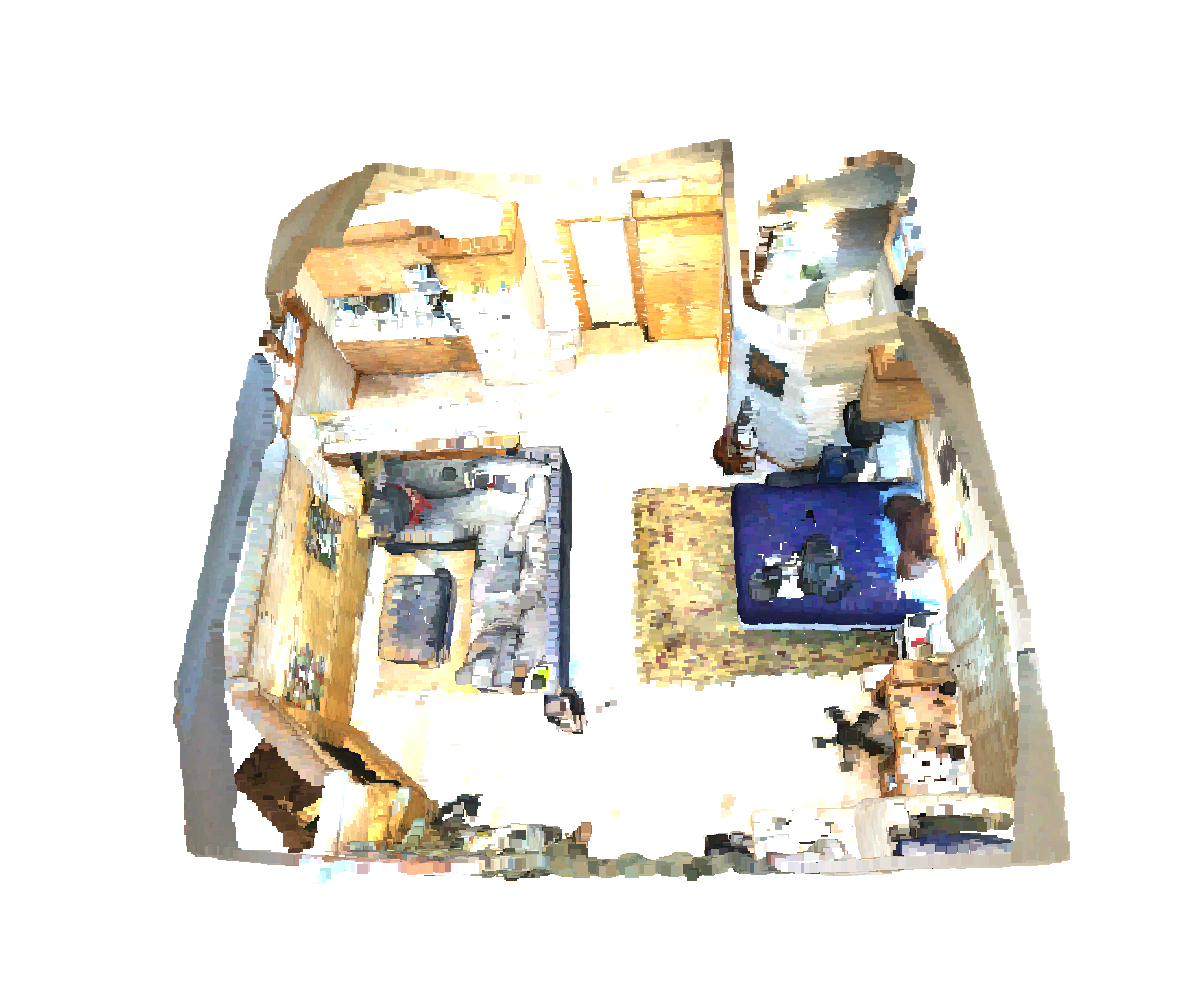}}
\end{minipage}
\caption{Scene representation using different numbers of points.}
\label{fig: fps}
\end{figure}

\begin{figure}[tb]
\begin{center}
  \includegraphics[width=1.0\linewidth]{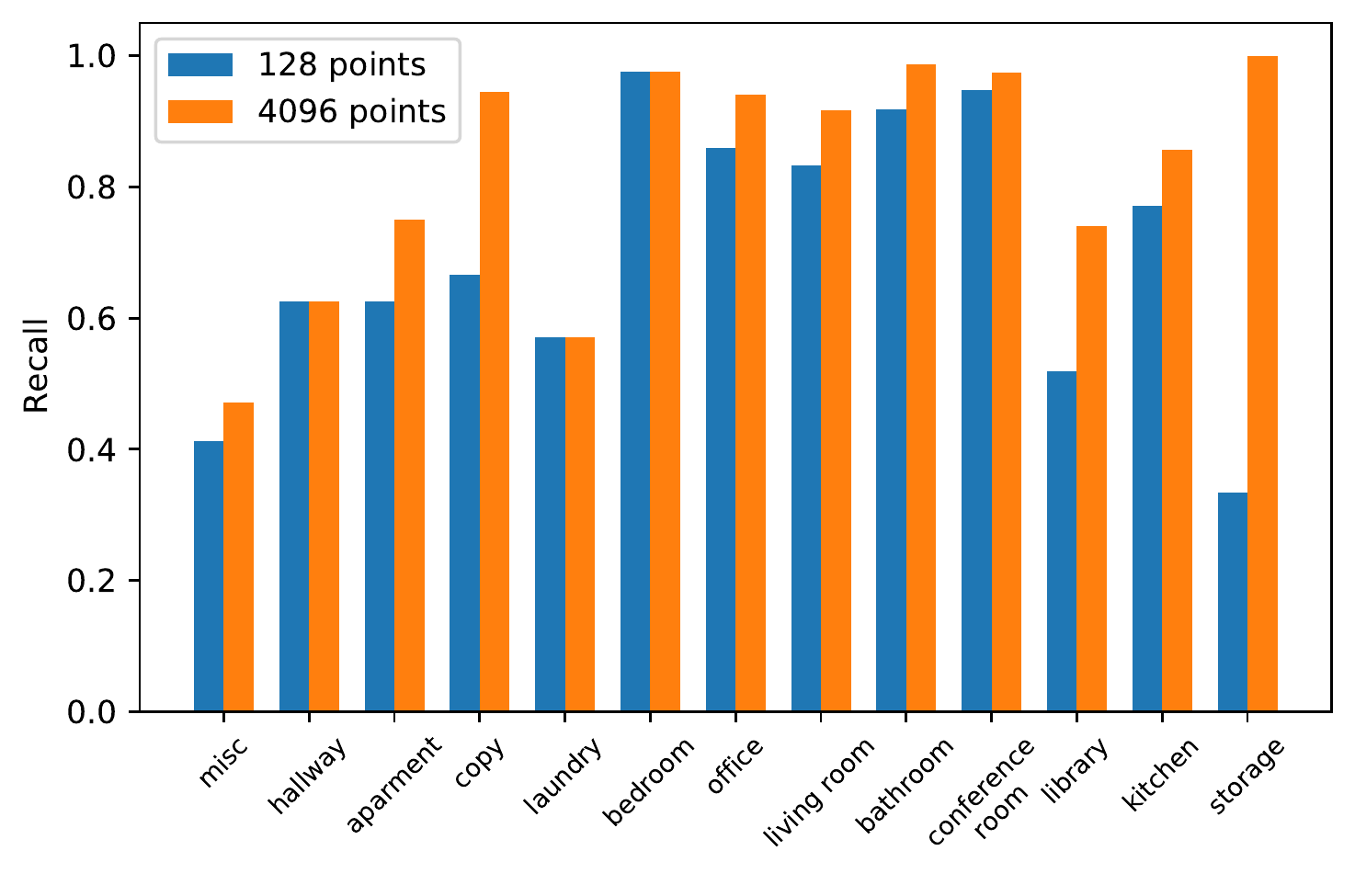}
\end{center}
  \caption{Recall per scene type with 128 and 4096 points as input.}
\label{fig: num_points_detail}
\end{figure}

The good performance of Pointnet-type methods, where the input is decimated to only 4096 points, raises the question how densely a scene must be sampled to successfully classify it.
We study the influence of point density by training and testing with different numbers of input points (without the semantic segmentation branch, since labeling individual points requires dense point clouds). In Fig.~\ref{fig: fps} we visually illustrate different point counts (for Pointnet++, we directly change the subsampling, for Resnet14 we first downsample, then voxelise). Results for scene type classification with different counts are shown in Fig.~\ref{fig:vary_points}. Unexpectedly, both Pointnet++ and Resnet14 achieve over 75\% accuracy even with only 128 points, and between 512 and 1024 points performance saturates at around 85\%.
The high accuracy with very sparse point clouds extends across most classes, larger drops are observed only for the rather rare \emph{storage} and \emph{copy room} classes, and to some degree for \emph{library}. See Fig.~\ref{fig: num_points_detail}.
To conclude, very sparse point clouds that are even difficult to interpret visually are sufficient to classify the scene type with high accuracy, with or without colour. 
In the context of robotics this is particularly interesting: a "rough glance" at the environment is apparently enough to determine the scene type in many cases, supporting the strategy to perform scene recognition early on and invoke scene-specific algorithms for detailed reconstruction or interaction. That sparse "first glance" could, for instance, come from SLAM keypoints, a low-resolution stereo matcher, or a fast overview scan.

\subsection{Geometry or semantics?}
We do point out that, while we have just shown that few points are enough for stand-alone scene recognition, an important limitation is that the performance can then no longer be boosted with multi-task learning.
Semantic segmentation necessitates high point density. This is nicely illustrated by the geometry-free Random Forest baseline. If we train the semantic segmentation network on downsampled point clouds with only 128 points, its object class predictions are wildly off, and scene type prediction based on their frequency histograms reaches only 42\% accuracy. While still far from chance level, this is clearly not a useful result.
On the contrary, predicting from 128 points with the true (oracle) object class labels reaches 81\% accuracy. Note, at so low point counts the geometry-free approach has an advantage, since fewer samples are needed to capture the frequency distribution than to capture the scene shape. 
In summary, geometric layout goes a long way and is effective even with very small, sparse point clouds. Multi-task learning with semantic segmentation as auxiliary task can improve scene type classification further, but is needs high point density. It remains to be explored whether there are other auxiliary tasks that can operate at low density.

\subsection{Colour}
\begin{figure}[tb]
\begin{center}
  \includegraphics[width=1.0\linewidth]{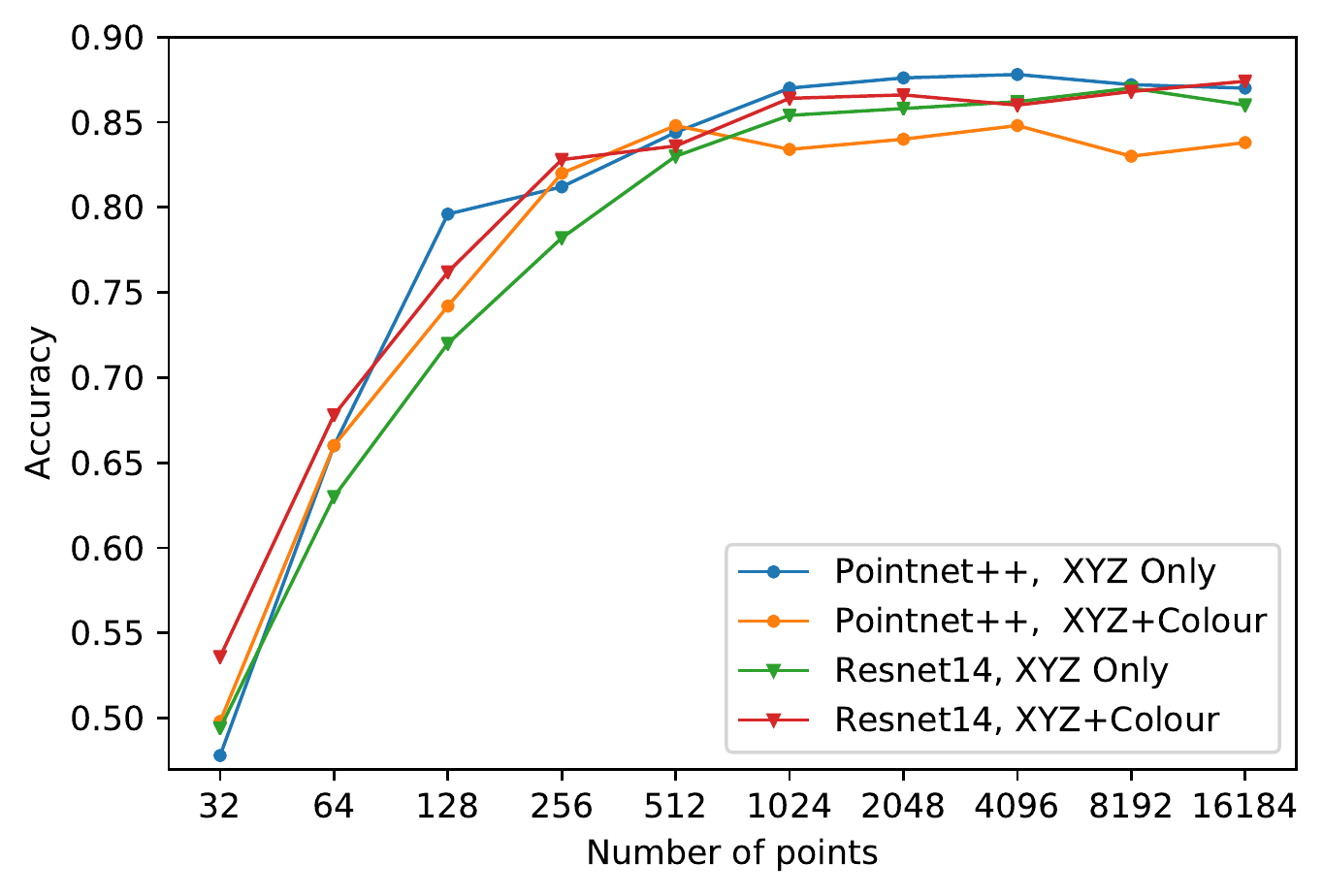}
\end{center}
\caption{Scene classification results for different point densities.}
\label{fig:vary_points}
\end{figure}

Colour information is not always available for 3D scan data. Table \ref{tab: comparison} shows the difference between models trained with and without colour information for the points. For the stronger models we also show the influence of colour at different point densities in Fig.~\ref{fig:vary_points}.
While already a global colour histogram greatly beats random chance, and low-performing models like vanilla Pointnet get a significant boost from colour, the effect on the best-performing models is small (but rather consistent) -- scene recognition does not seem to critically depend on colour.
At least on ScanNet, colour information degrades the performance of Pointnet++, while it slightly improves Resnet14.
A caveat is again that despite its limited impact, eschewing colour when it would be available may weaken multi-task learning, as semantic segmentation often does benefit quite a lot from colour.

\subsection{Influence of individual object classes}
\begin{figure}[tb]
\begin{center}
  \includegraphics[width=1.0\linewidth]{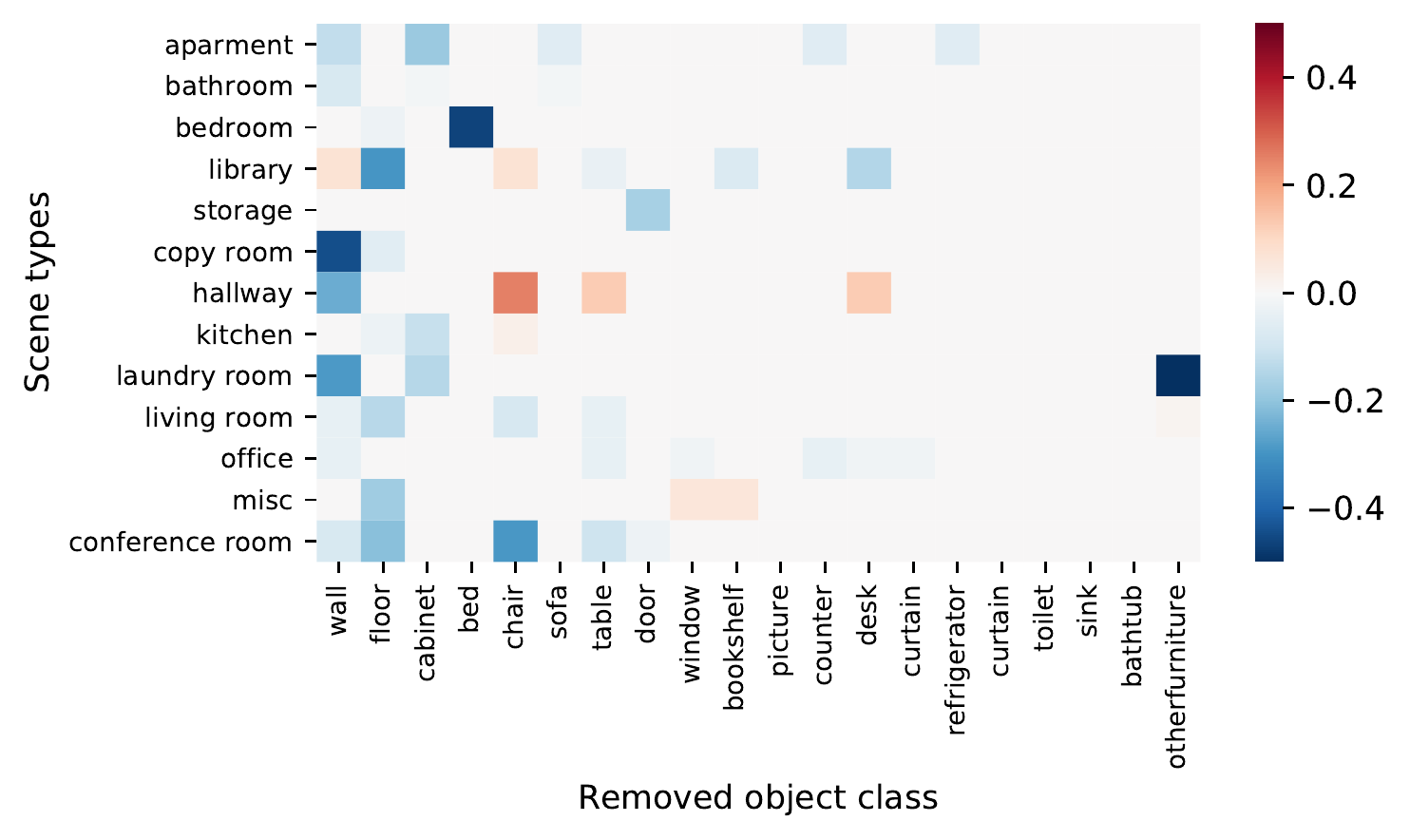}
\end{center}
  \caption{Change of recall per scene type after removing all points with a certain object class label.}
\label{fig:remove_semantics}
\end{figure}

Intuitively, one might suspect that individual "marker" objects are enough to determine the scene type, e.g., bathtubs appear only in bathrooms, beds point to bedrooms or apartments, bookshelves to libraries, etc. 
To check to what degree scene recognition degenerates to detecting individual objects, we remove the points belonging to each object class in turn from the test data (without changing the model trained on complete scenes).
In Fig.~\ref{fig:remove_semantics} we show the difference in recall for each scene type if removing points of a certain class. The average magnitude of the changes is only 2 percent points, with  few exceptions the absence of an individual object class changes the recall by $<$5 pp.

There are a few examples of marker objects -- we observe pronounced drops for \emph{bedroom} when removing the beds, and for \emph{laundry room} when removing \emph{unknown furniture} (likely due to some frequent object class that was not labeled separately, such as laundry baskets). On the contrary the \emph{hallway} class benefits from removing chairs, apparently this prevents confusions with other scene types.
Coming back to a seemingly obvious example that we have mentioned before, removing the bathtub(s) does not change the recall of bathrooms at all. In contrast, almost all classes suffer at least a bit from removing walls or floors, again pointing to the role of overall scene shape.
While the presence of certain object types certainly plays a role, scene type classification is not only the search for a specific marker object, but appears to exploit more complex patterns such as co-occurrence and spatial layout of objects.

\subsection{Completeness}
\begin{figure}[tb]
\begin{center}
  \includegraphics[width=1.0\linewidth]{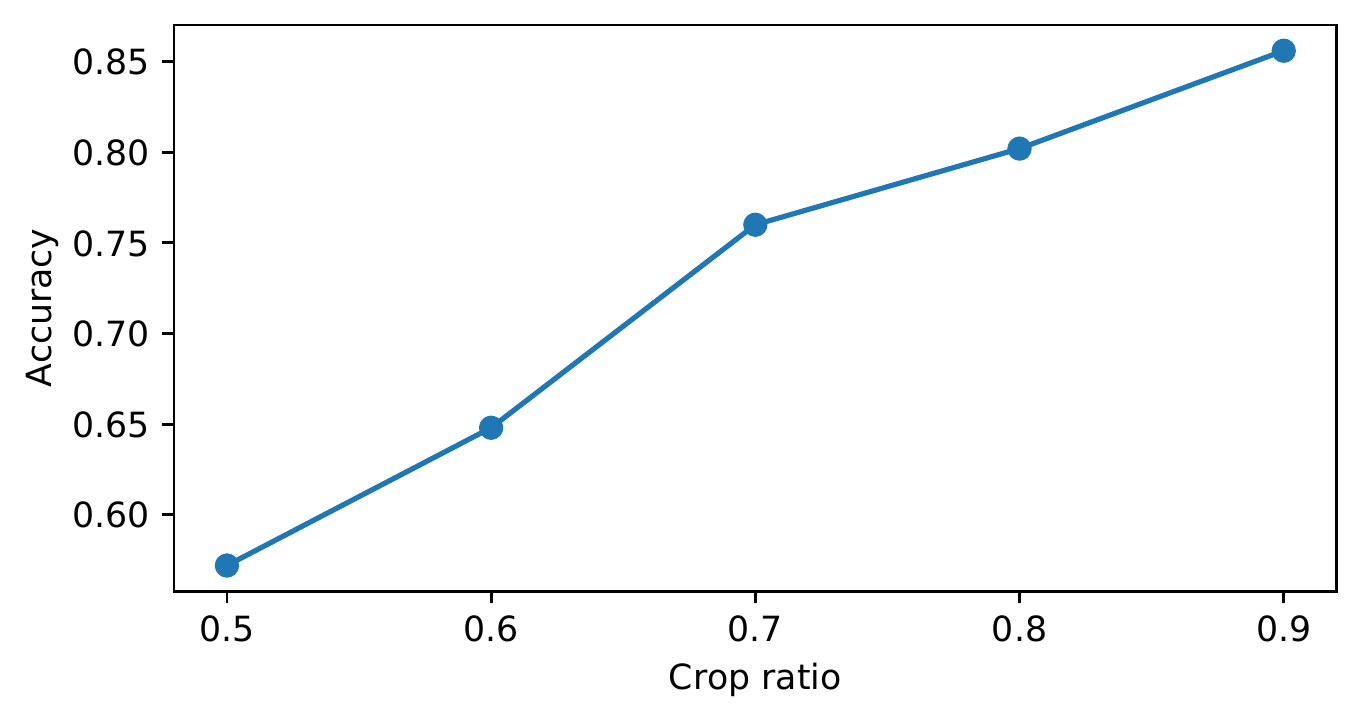}
\end{center}
  \caption{Scene classification results w.r.t. the crop ratio of the scene.}
\label{fig:crop_scenes}
\end{figure}

Mobile robots often cannot capture the complete scene due to occlusions or non-panoramic sensors.
An overall exploration that takes into account sensor-specific or application-specific occlusion patterns is beyond the scope of this paper, but we study tolerance to incomplete scenes with a simple cropping scheme: we cut out an axis-aligned corner of the scene bounding box with varying crop ratio, where the latter is defined as the fraction of the $x$- and $y$-axes that is retained (so, e.g., a crop ratio of 0.7 means that we retain a box that spans 70\% of the bounding box in $x$- and $y$-direction, over the entire height of the scene).
As can be seen in Fig.~\ref{fig:crop_scenes}, accuracy drops to $<$58\% at a crop ratio 0.5, corresponding to a scene of which only $\frac{1}{4}$ has been observed, and grows approximately linearly to reach $>$85\% for the complete scene.
In other words, there is some robustness, e.g., at crop ratio 0.8 (keeping $\approx\frac{2}{3}$ of the original scene) the performance penalty is $<$5\%.
But overall, completeness matters. When larger coherent portions of the scene are missing the classification error increases fairly quickly. It is preferable to capture a coarse, but complete 3D view of the room rather than a detailed, but partial one.
This further supports our case for working in 3D rather than classify based on 2D images or 2.5D range images.
Note that we again only modified the test scenes. While the cutout regularisation used in our training is somewhat similar to the described cropping, it may be possible to improve robustness against incomplete scenes with more sophisticated data augmentation schemes.

\subsection{Hard samples} 
\begin{figure}[tb]
\begin{center}
  \includegraphics[width=1.0\linewidth]{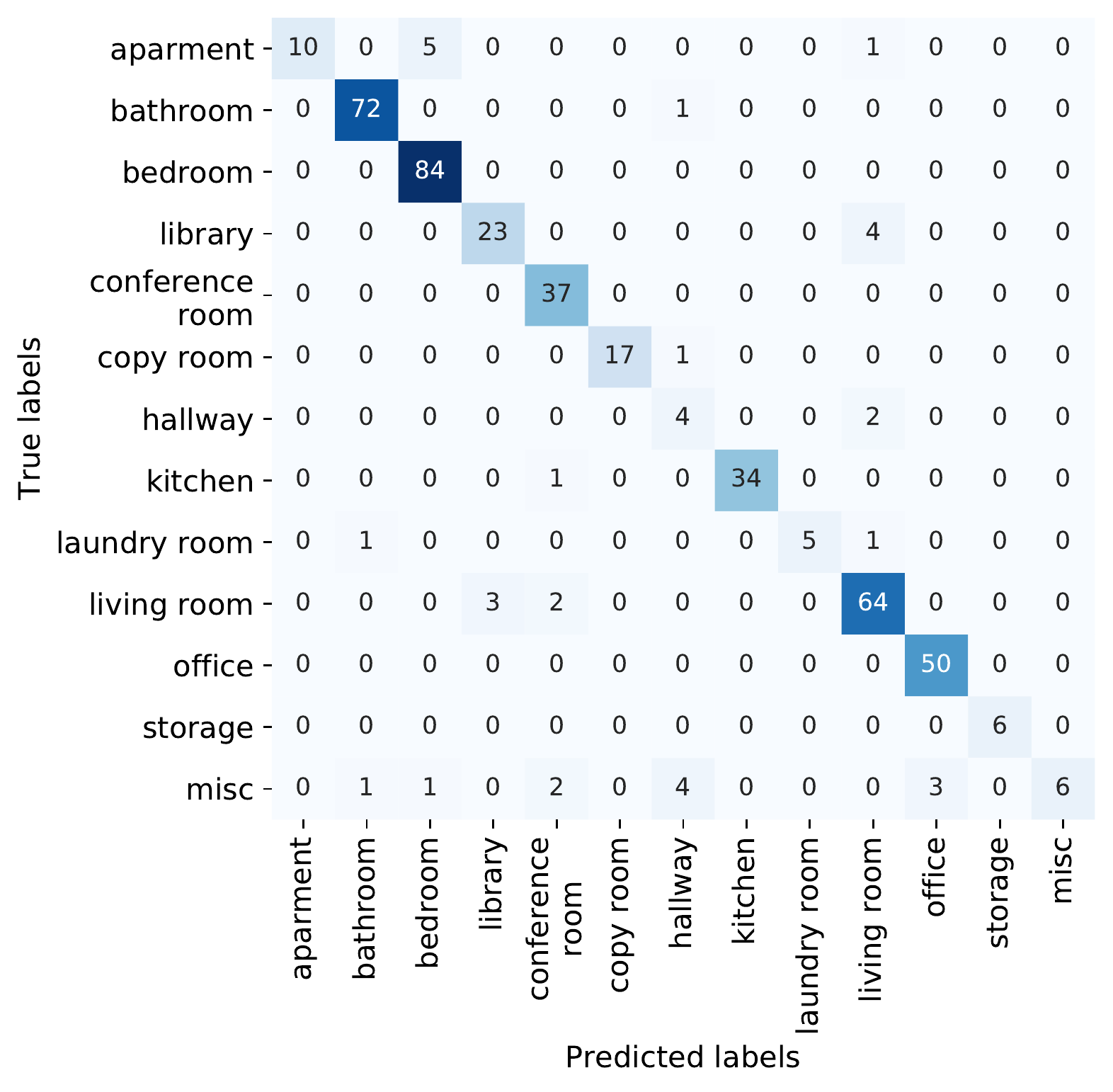}
\end{center}
  \caption{Confusion matrix calculated on validation dataset.}
\label{fig: cm}
\end{figure}

\begin{figure*}[tb]
\begin{minipage}{1\linewidth}
\centering
\subfloat{\includegraphics[width=0.25\linewidth]{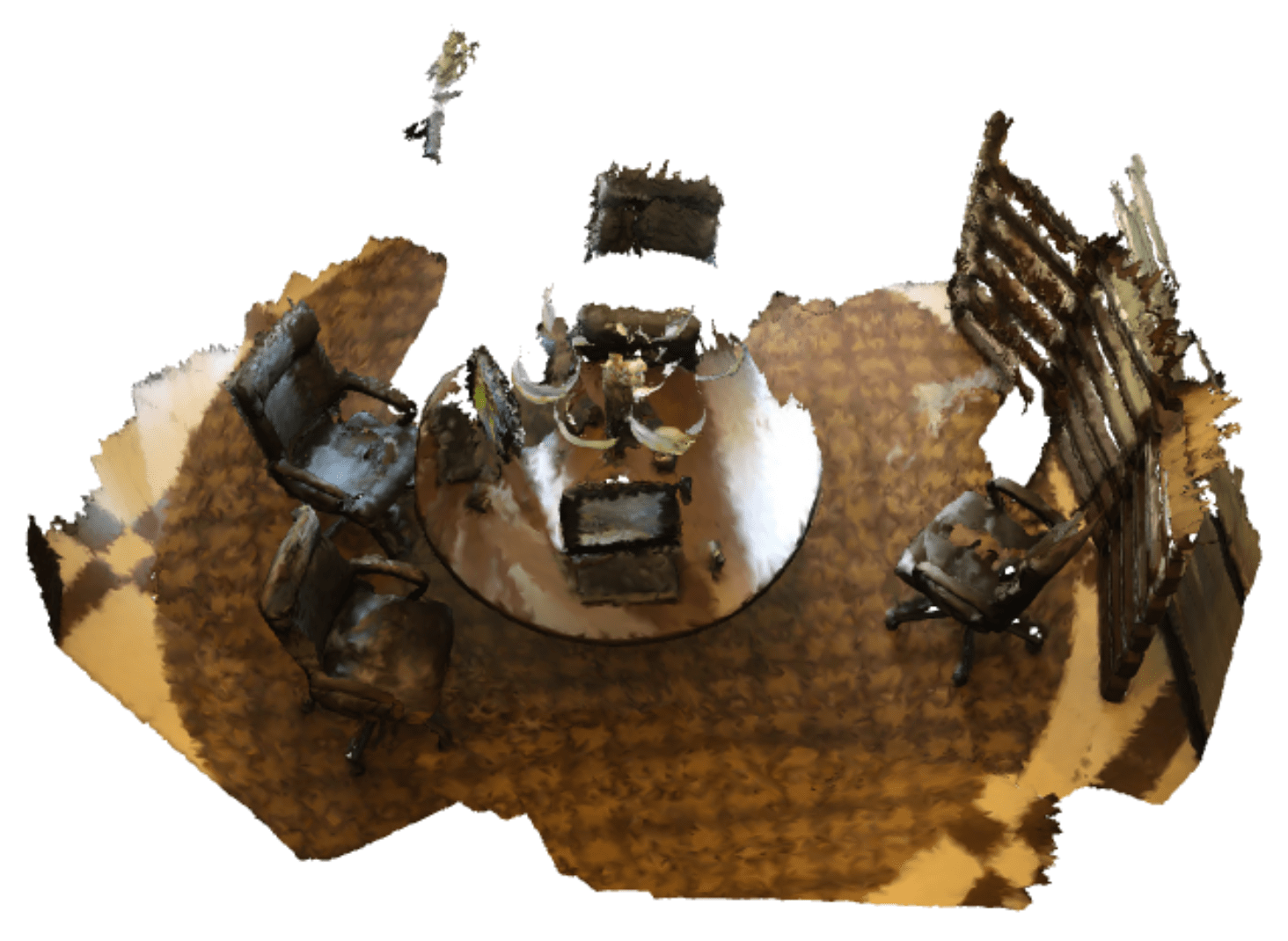}}
\subfloat{\includegraphics[width=0.25\linewidth]{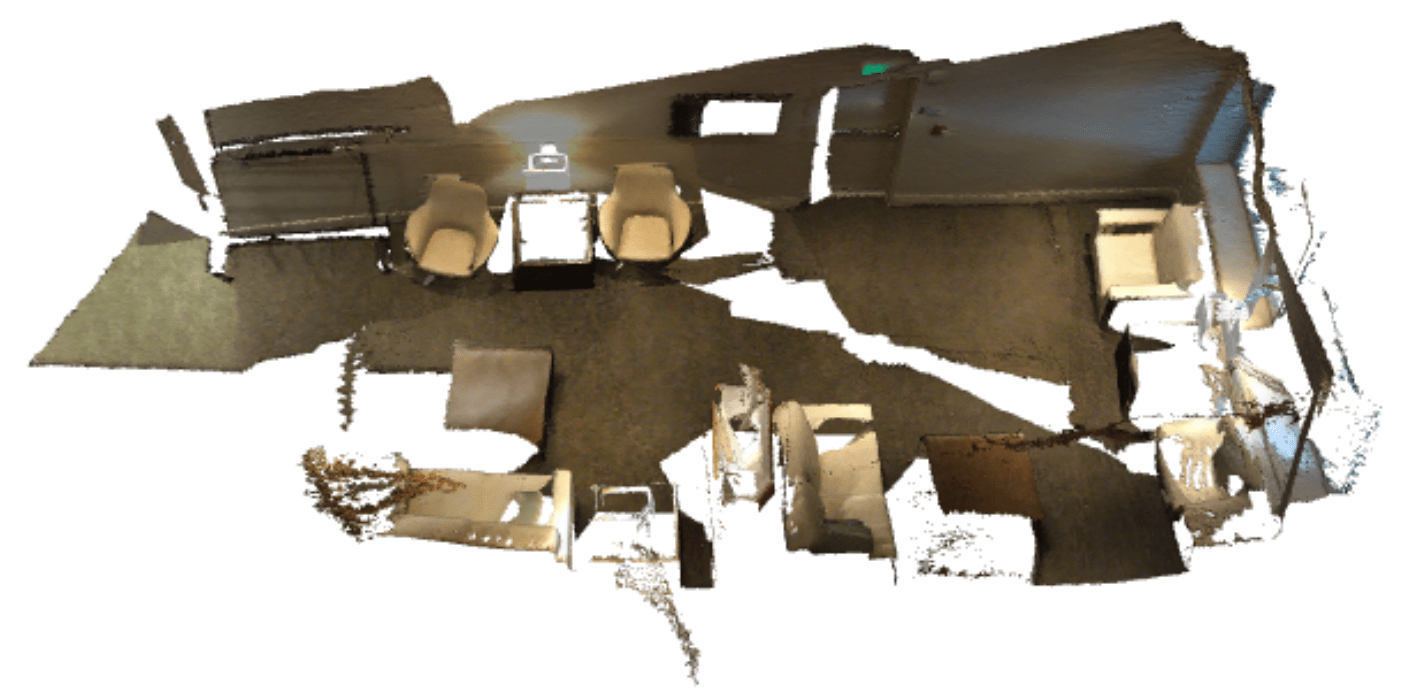}}
\subfloat{\includegraphics[width=0.25\linewidth]{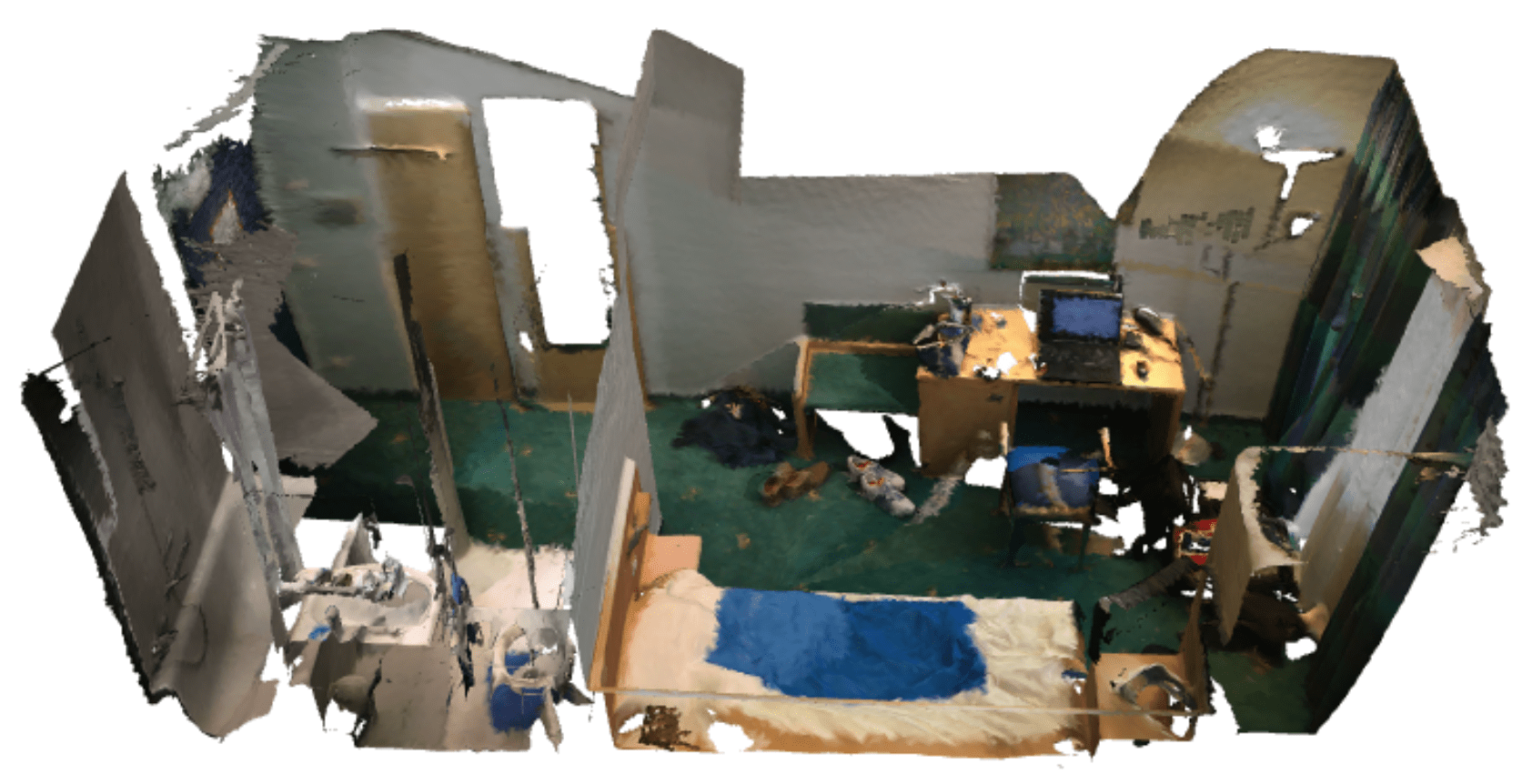}}
\subfloat{\includegraphics[width=0.25\linewidth]{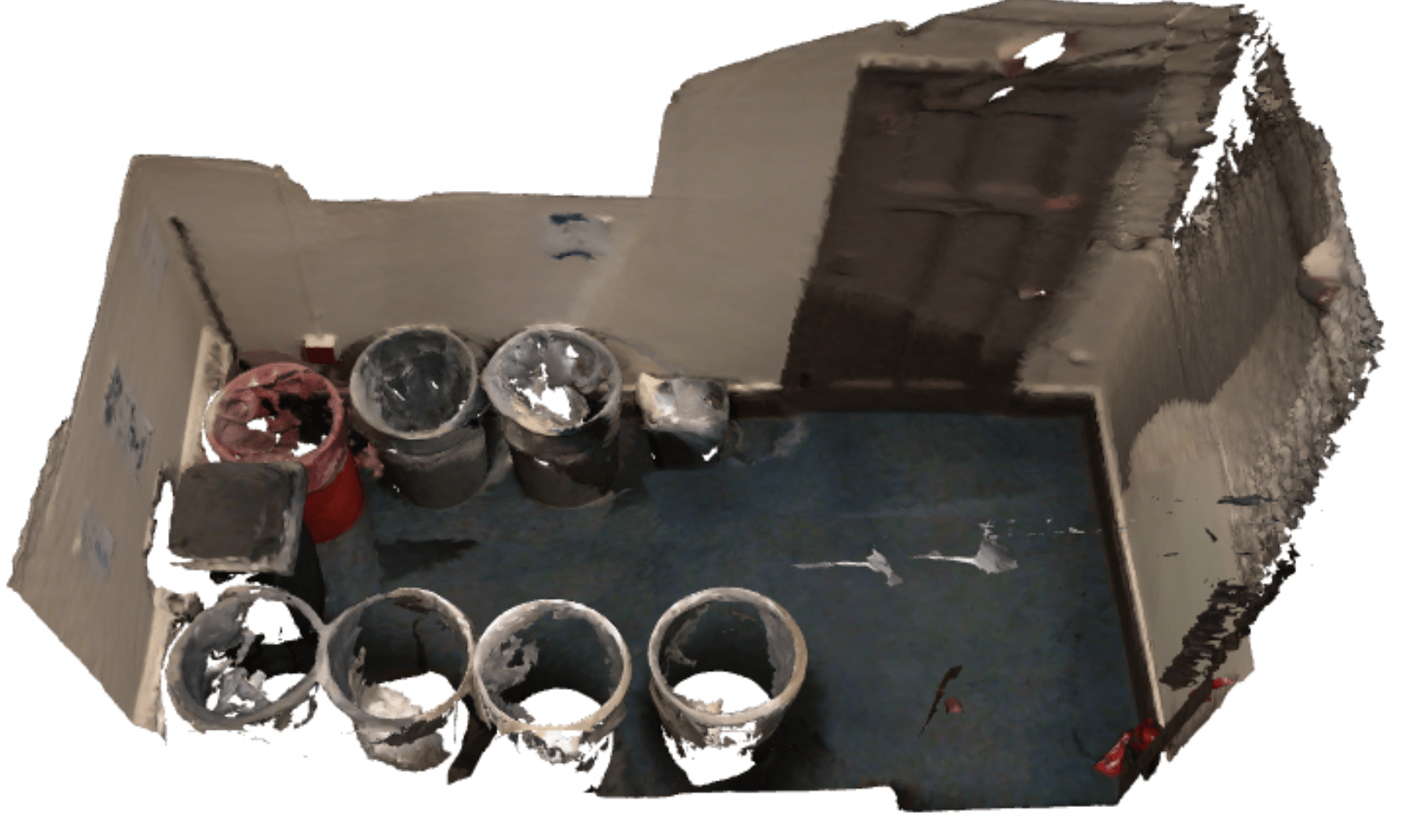}}
\end{minipage}

% \begin{minipage}{1\linewidth}
% \centering
% \subfloat{\includegraphics[width=0.4\linewidth,height=2.5cm]{figures/lobby-living-min.png}}
% \subfloat{\includegraphics[width=0.6\linewidth]{figures/mis_4.pdf}}
% \end{minipage}

\begin{minipage}{1\linewidth}
\centering
\subfloat{\includegraphics[width=0.25\linewidth]{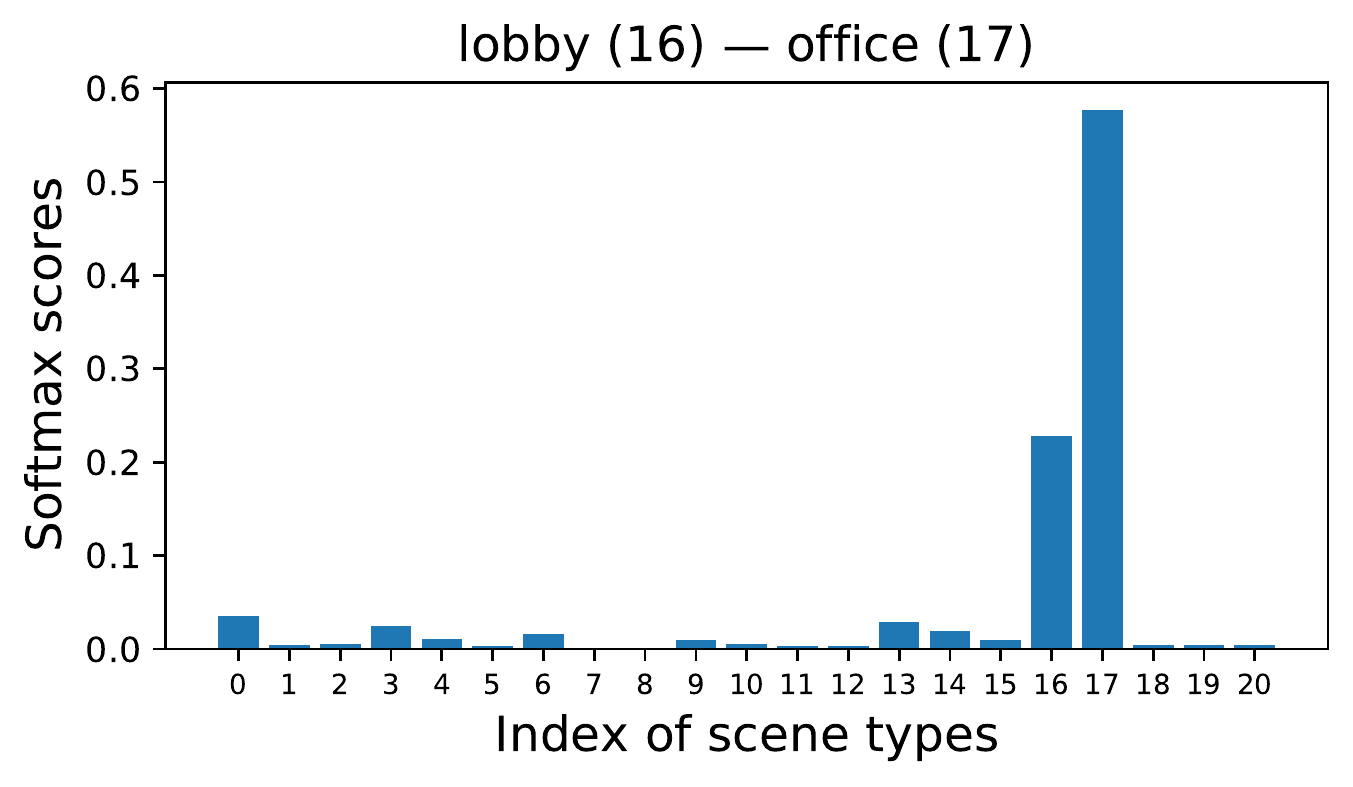}}
\subfloat{\includegraphics[width=0.25\linewidth]{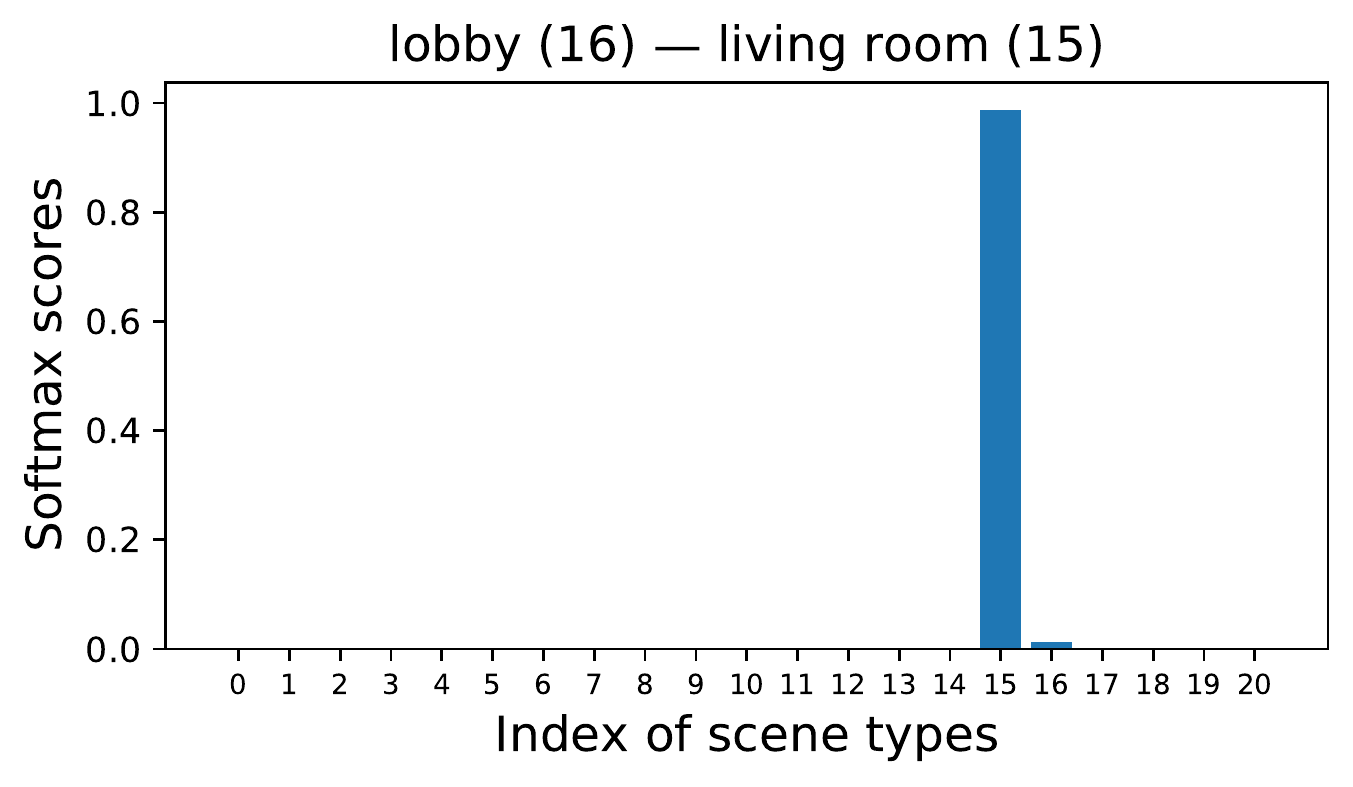}}
\subfloat{\includegraphics[width=0.25\linewidth]{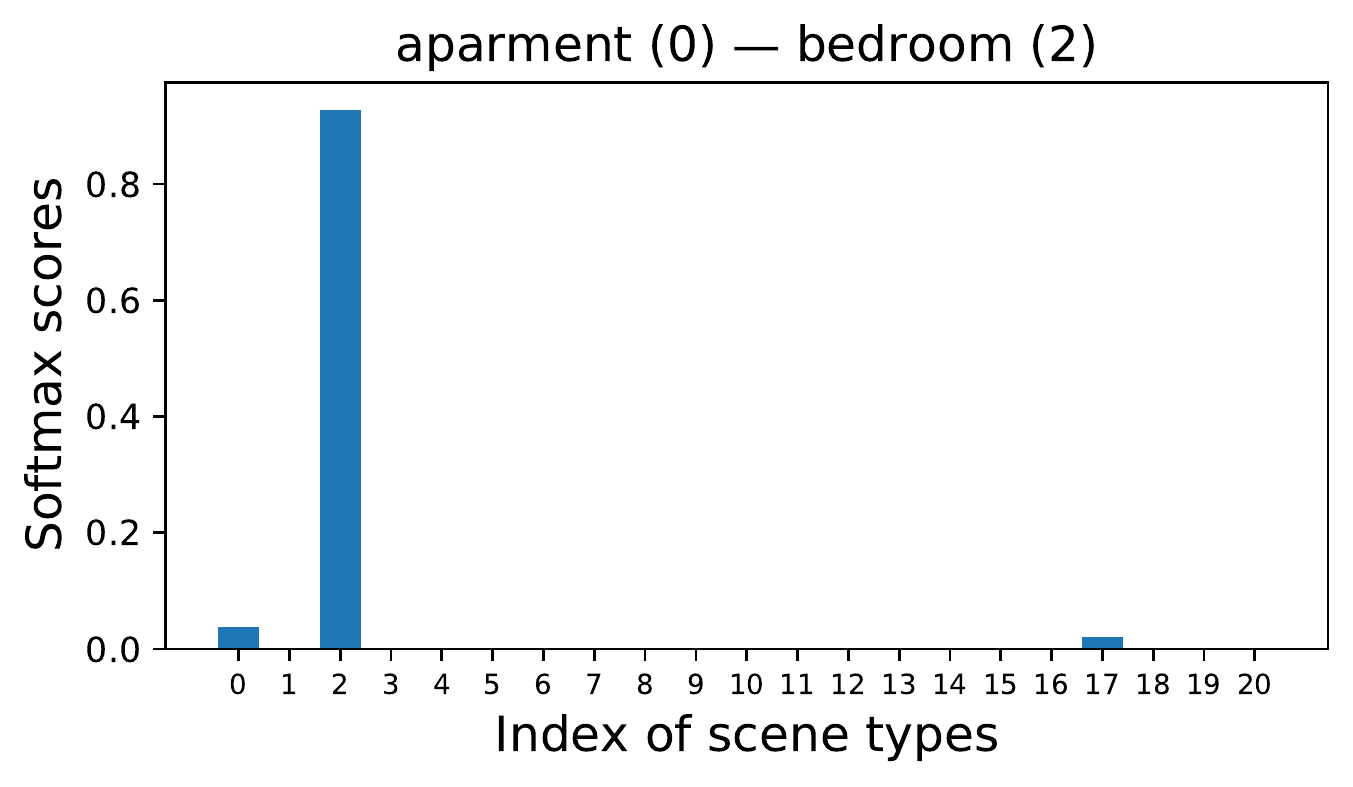}}
\subfloat{\includegraphics[width=0.25\linewidth]{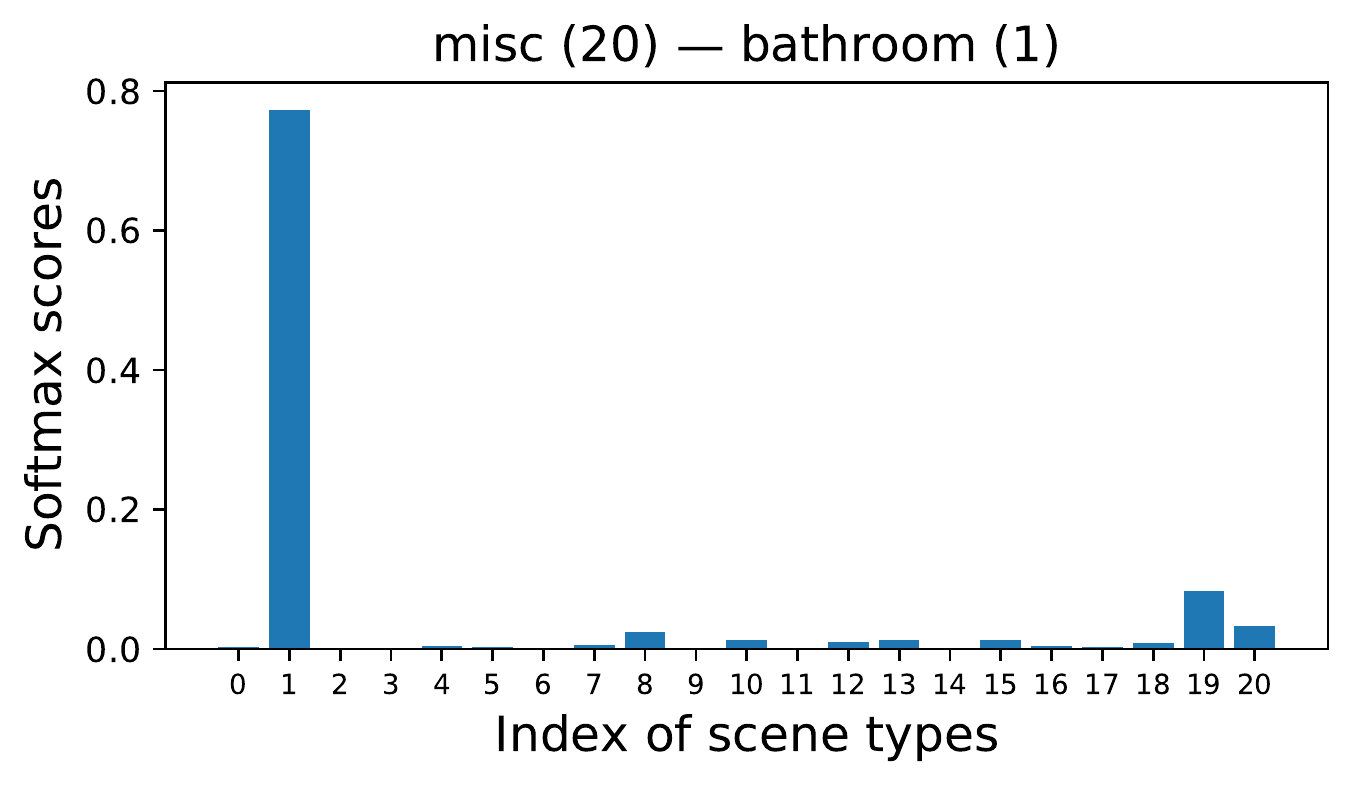}}
\end{minipage}

% \begin{minipage}{1\linewidth}
% \centering
% \subfloat{\includegraphics[width=0.4\linewidth,height=2.5cm]{figures/misc-bathroom-min.png}}
% \subfloat{\includegraphics[width=0.6\linewidth]{figures/mis_8.pdf}}
% \end{minipage}

\caption{Miss-classified samples. The first row is the scene. The second row shows the class scores, true and predicted scene types.}
\label{fig: hard-samples}
\end{figure*}

In total, we have 500 validation samples. 50 of them cannot be classified correctly by our best setup. From the confusion matrix in Fig.~\ref{fig: cm} we can see that most of the miss-classified scenes are from the classes
\emph{apartment} (often mistaken for \emph{bedroom}), \emph{living room} and \emph{misc}. The latter is an open rejection class for scenes of undefined type or whose reconstruction failed, thus difficult to learn. Moreover, the miss-classifications are often visually plausible, and several of them are "near misses" where the correct class has the second-highest score (see
Fig.~\ref{fig: hard-samples}).

\section{Conclusion}
We have, to our knowledge for the first time, performed a systematic study of indoor scene recognition with 3D scene representations (point clouds or voxels). We found that working in 3D greatly improves scene recognition. It turns out that there are two different cues which both achieve high accuracy: on the one hand the global scene geometry alone reaches up to 87\% accuracy on ScanNet, on the other hand also the distribution of semantic object labels by itself reaches 82.8\%.  Combining the two cues in a multi-task learning framework boosts the accuracy to 90.3\%.
This suggests two main modes of operation: When classifying purely based on geometry, only a coarse representation with few points is sufficient, making it possible to achieve a very reasonable accuracy quickly and with little data.
On the other hand, when aiming for the best possible accuracy, it is beneficial to rely on a dense sampling of the scene and per-point labels for the training data, so as to leverage point-wise semantic segmentation as auxiliary task. 
While our results support the use of 3D representations, it remains an open question whether one must go all the way to irregular point clouds or voxels. It may be interesting to explore alternative representations that capture the global scene structure in an efficient 2D projection, such as for instance (sparse?) depth panoramas.

\bibliography{IEEEabrv,IEEEtran}
\bibliographystyle{IEEEtran}

\end{document}